%% file: neurips_2025.tex
\documentclass{article}

\PassOptionsToPackage{numbers, compress}{natbib}

\usepackage[preprint]{neurips_2025}

\usepackage{microtype}
\usepackage{graphicx}
\usepackage{subfigure}
\usepackage{booktabs} 
\usepackage{multirow}
\usepackage{xcolor}
\usepackage{tcolorbox}
\usepackage{colortbl}

\usepackage[utf8]{inputenc} 
\usepackage[T1]{fontenc}    
\usepackage{hyperref}       
\usepackage{url}            
\usepackage{booktabs}       
\usepackage{amsfonts}       
\usepackage{nicefrac}       
\usepackage{microtype}      
\usepackage{xcolor}         

\usepackage{amsmath}
\usepackage{amssymb}
\usepackage{mathtools}
\usepackage{amsthm}
\theoremstyle{definition} 
\usepackage{enumitem} 

\tcbuselibrary{theorems}
\usepackage{amsfonts, mathrsfs}
\usepackage{makecell} 
\usepackage{multirow} 
\usepackage{wrapfig}
\usepackage[numbers]{natbib}
\definecolor{drop_green}{RGB}{0,100,0}
\definecolor{bonus_red}{RGB}{180,0,0}
\definecolor{bg}{RGB}{176,226,255}

\newcommand{\bbonus}[1]{{\textcolor{bonus_red}{$\uparrow{#1}$}}}
\newcommand{\bonus}[1]{{\textcolor{drop_green}{$_{\uparrow#1}$}}}
\newcommand{\ddrop}[1]{{\textcolor{drop_green}{$\downarrow{#1}$}}}
\newcommand{\drop}[1]{{\textcolor{bonus_red}{$_{\downarrow#1}$}}}

\usepackage[capitalize,noabbrev]{cleveref}

\title{How Far Are We from Optimal Reasoning Efficiency?}

\newcommand{\method}{REO-RL}
\newcommand{\methodname}{Reasoning Efficiency Optimization with Reinforcement Learning}

\newcommand{\gjx}[1]{{\textcolor{black}{#1}}}

\newtcbtheorem[number within=section]{mydefinition}{Definition}{
    colback=red!5!white,
    colframe=red!75!black,
    fonttitle=\bfseries,
    width=\textwidth
}{def}

\newtcbtheorem[number within=section]{mymethod}{Method}{
    colback=teal!5!white,
    colframe=teal!70!black,
    fonttitle=\bfseries,
    width=\textwidth
}{def}

\author{Jiaxuan Gao$^{1}$
	\quad
	 Shu Yan$^{3}$ \quad Qixin Tan$^{1}$ \quad Lu Yang$^{1}$ \\
\textbf{Shusheng Xu$^{12} \quad $Wei Fu$^{12}$ \quad Zhiyu Mei$^{12}$ \quad Kaifeng Lyu$^{1}$ \quad Yi Wu$^{1}$\thanks{~Corresponding author}
} \\
$^1$ IIIS, Tsinghua University \quad $^2$ Ant Research, RL Lab \quad
$^3$ Nanjing University\\
\texttt{\{samjia2000, jxwuyi\}@gmail.com} \\
}

\begin{document}

\maketitle

\input{sections/00_abs}

\input{sections/10_intro}

\input{sections/20_related}

\input{sections/30_prelim}

\input{sections/40_length_scaling}

\input{sections/60_experiments}

\input{sections/70_conclusion}

\newpage

{
\bibliographystyle{plainnat}
\bibliography{reference}
}

\newpage

\appendix
\input{sections/appendix}

\newpage
\clearpage 
\newpage

\end{document}

%% file: sections/00_abs.tex
\begin{abstract}
Large Reasoning Models (LRMs) demonstrate remarkable problem-solving capabilities through extended Chain-of-Thought (CoT) reasoning but often produce excessively verbose and redundant reasoning traces. This inefficiency incurs high inference costs and limits practical deployment. While existing fine-tuning methods aim to improve reasoning efficiency, assessing their efficiency gains remains challenging due to inconsistent evaluations. In this work, we introduce the \emph{reasoning efficiency frontiers}, empirical upper bounds derived from fine-tuning base LRMs (DeepSeek-R1-Distill-Qwen-1.5B/7B \gjx{and Qwen3-4B/8B}) across diverse approaches and training configurations. Based on these frontiers, we propose the \emph{Reasoning Efficiency Gap (REG)}, a unified metric quantifying deviations of any fine-tuned LRMs from these frontiers. Systematic evaluation on challenging mathematical benchmarks reveals significant gaps in current methods: they either sacrifice accuracy for short length or \gjx{still remain inefficient under tight token budgets.} 
To reduce the efficiency gap, we propose {\method}, a \gjx{class of Reinforcement Learning algorithms} that \gjx{minimizes REG} by targeting a sparse set of token budgets. Leveraging numerical integration over strategically selected budgets, {\method} approximates the full efficiency objective with low error using a small set of token budgets. 
Through systematic benchmarking, we demonstrate that our new efficiency metric, REG, effectively captures the accuracy-length trade-off, with low-REG methods reducing length while maintaining accuracy. Our approach, {\method}, significantly improves reasoning efficiency, reducing REG by $\ge50\%$ across all evaluated LRMs and matching Qwen3-4B/8B efficiency frontiers under a 16K token budget with minimal accuracy loss. Ablation studies confirm the effectiveness of our exponential token budget strategy. Finally, our findings highlight that fine-tuning LRMs to perfectly align with the efficiency frontiers remains an open challenge.
We will release the code, data, and models.\footnote{\href{https://github.com/samjia2000/Optimal-Reasoning-Efficiency}{https://github.com/samjia2000/Optimal-Reasoning-Efficiency}}.
\end{abstract}

%% file: sections/10_intro.tex
\section{Introduction}

Large Reasoning Models (LRMs) have recently emerged as a powerful class of models capable of solving complex tasks that require advanced reasoning. Frontier LRMs such as OpenAI o1~\citep{openai_learning_to_reason} and DeepSeek R1~\citep{guo2025deepseek} have obtained superior performance across a wide range of tasks, including mathematical reasoning and competitive programming. A major factor behind this success is their ability to perform deep, multi-step reasoning through extended Chain-of-Thought (CoT) processes. These reasoning traces often include sophisticated operations such as reflection, verification, and exploration, within a single inference pass. 

However, the powerful capability of long CoT reasoning comes at a cost. LRMs frequently generate overly verbose and redundant reasoning traces, a phenomenon referred to as the \emph{overthinking problem}~\citep{yang2025thinkingoptimalscalingtesttimecompute,sui2025stop}.
Recent studies~\citep{chen2024not,sui2025stop} have shown that even simple questions like “2 + 3 = ?” can result in outputs spanning up to 900 tokens. This redundancy brings a significant cost in inference time and limits practical deployment. 
Several fine-tuning approaches have been proposed to improve reasoning efficiency, \gjx{with representative examples being length reduction with RL}~\citep{team2025kimi, luo2025o1, aggarwal2025l1, arora2025training, yeo2025demystifying, shen2025dast, qu2025optimizing, yang2025thinkneedselfadaptivechainofthought, she2025hawkeyeefficientreasoningmodelcollaboration, hou2025thinkprune}, \gjx{and hybrid reasoning approaches that train LRMs to adaptively select thinking or no-thinking modes~\citep{jiang2025thinkneedlargehybridreasoning, lou2025adacotparetooptimaladaptivechainofthought, zhang2025adaptthinkreasoningmodelslearn, liang2025thinkswitcherthinkhardthink}}. However, comparing these methods remains difficult due to inconsistent experiment setups, including varying models, evaluation benchmarks, and mixed performance metrics. It is still unclear how close current approaches are to the optimal trade-off between length and accuracy.

\gjx{In this work, we investigate a critical question:} {\emph{How far are the current approaches from reaching the optimal reasoning efficiency?}}
To answer this question, we conduct a comprehensive empirical study using a set of LRMs, \gjx{DeepSeek-Distill-Qwen-1.5B/7B and Qwen3-4B/8B}, on a set of challenging mathematical reasoning benchmarks. We introduce the concept of \emph{\textbf{reasoning efficiency frontiers}}, derived from fine-tuning the base LRMs with \gjx{various} types of algorithms and diverse training configurations.
These reasoning efficiency frontiers represent the best reward achievable by the current approaches at each token budget, offering a practical lower bound on optimal efficiency. 
By comparing current methods to these frontiers, we uncover a substantial gap. Existing methods often fall short in one of two ways, either \emph{(1) they aggressively shorten responses at the expense of accuracy}, or \emph{(2) methods that reach high overall accuracy would consume significantly more tokens than necessary to reach moderate accuracy levels}. 
To quantify this gap, we propose the \emph{\textbf{Reasoning Efficiency Gap (REG)}}, a unified metric that captures both accuracy and response length by measuring the area between the budget-accuracy curve of an LRM and the \emph{efficiency frontiers}. \gjx{Specifically, REG is computed as the accuracy gap between the efficiency frontiers and the LRM across all token budgets. REG also quantifies the average amount of wasted tokens across all accuracies.} More importantly, REG offers practical insights into how much room still remains for improvement.

We further ask: \emph{How can an LRM be fine-tuned to minimize this efficiency gap?} 
A natural approach is to optimize the rewards across all possible token budgets at RL training time. 
However, this approach leads to a costly training process since rewards across all token budgets should be evaluated. 
To overcome the inefficiency of this dense reward approach, we introduce \textbf{\emph{{\methodname} ({\method})}}, \gjx{a novel family of RL algorithms for improving reasoning efficiency of LRMs.}
The key insight of {\method} is that the total rewards across all token budgets can be well-approximated using numerical integration over a small set of representative token budgets. 
\gjx{We introduce two installations of {\method}, including {\method} (Oracle) that uses an oracle-based greedy approach to select token budgets based on the estimated reasoning efficiency frontiers, {\method} (Exp) that selects exponentially spaced token budgets.}

\gjx{Through systematically benchmarking a diverse set of existing methods, we find that our efficiency metric REG effectively captures the trade-off between accuracy and response length. Methods that achieve low REG could significantly reduce the average response length with slight or even no accuracy drop. By reducing the efficiency gap, our approach, {\method}, outperforms baselines in terms of reasoning efficiency, and reduces REG by at least 50\% across all evaluated LRMs. Notably, {\method} matches the efficiency frontiers of Qwen3-4B/8B under 16K token budget, with only slight overall accuracy drop.}
Our ablation study further validates the success of the exponential token budget selection strategy and effective approximation with a small amount of token budgets.
\gjx{Finally, our findings highlight an unresolved challenge: \emph{fine-tuning LRMs to precisely align with their efficiency frontiers remains an open problem.}}

%% file: sections/20_related.tex
\section{Related Works}

\paragraph{Efficient Reasoning.} 
Prior studies have shown that LRMs often suffer from redundant reasoning. Even for very simple questions, Frontier LRMs often generate lengthy responses spanning thousands of tokens~\citep{chen2025think23overthinkingo1like,sui2025stop}. This redundancy in the reasoning process brings significant overheads in the inference costs. Several works are then proposed to make LRM reasoning more concise.  \cite{team2025kimi, luo2025o1, aggarwal2025l1, arora2025training, yeo2025demystifying, shen2025dast, qu2025optimizing, yang2025thinkneedselfadaptivechainofthought, she2025hawkeyeefficientreasoningmodelcollaboration, hou2025thinkprune,qi2025optimizinganytimereasoningbudget} investigate RL training with length reward designs, mostly focusing on reducing the reasoning lengths. A line of works apply SFT to fine-tune LRMs on datasets with variable-length reasoning traces to elicit concise reasoning~\citep{yu2024distilling, wang2023self, han2024token} or adjustable length control~\citep{kang2024c3ot,xia2025tokenskip,ma2025cot,liu2024can,yu2025z1efficienttesttimescaling}. \gjx{Some recent works investigate training LRMs to adaptively select whether to perform an explicit thinking process or directly produce the final solution according to the input query~\citep{zhang2025adaptthinkreasoningmodelslearn,tu2025learningthinkshapingadaptive,jiang2025thinkneedlargehybridreasoning}.} 
\gjx{Test-time techniques are also applied for enhancing} the reasoning efficiency through reward model guided decoding~\citep{sun2024fast, liao2025reward}, uncertainty-based dynamic reasoning~\citep{fu2024efficiently, fu2025reasoning}, and confidence-based approaches~\citep{taubenfeld2025confidence, huang2025efficient}. In this work, we study the optimal reasoning efficiency for LRMs and focus on training-based approaches for enhancing LRM reasoning efficiency. Our method aims at enhancing the accuracy of the LRM under diverse token budgets without explicitly incentivizing shorter responses.

\paragraph{RL for LRM Reasoning.} Reinforcement Learning is the central technique for eliciting and enhancing the reasoning capability of LRMs. Frontier LRMs, including OpenAI o1~\citep{openai_learning_to_reason} and DeepSeek R1~\citep{guo2025deepseek}, have shown that applying "zero RL" on a base LLM could effectively elicit the ability to utilize long CoTs for complex reasoning. A series of works have emerged with the focus on improving the training efficiency of RL for LRMs from the perspectives of data~\citep{deepscaler2025, areal2025, skywork-or1-2025, li2025limrrlscaling, wang2025reinforcementlearningreasoninglarge}, algorithms~\citep{guo2025deepseek,skywork-or1-2025,deepscaler2025, yu2025dapoopensourcellmreinforcement,yue2025vapoefficientreliablereinforcement}, and training framework~\citep{sheng2024hybridflow,deepcoder2025,areal2025}. A number of works successfully apply zero RL training on a wide range of reasoning-heavy domains, including multi-modality~\citep{shen2025vlm,zhang2025r1}, medical~\citep{yu2025finemedlmo1enhancingmedicalreasoning,chen2024huatuogpto1medicalcomplexreasoning}, and financial~\citep{liu2025finr1largelanguagemodel}. Recent works also explore efficiency enhancement by encouraging concise reasoning with RL~\citep{team2025kimi, luo2025o1, aggarwal2025l1, arora2025training, yeo2025demystifying, shen2025dast, qu2025optimizing, yang2025thinkneedselfadaptivechainofthought, she2025hawkeyeefficientreasoningmodelcollaboration, hou2025thinkprune}. In this work, we focus on enhancing the reasoning efficiency with RL.

%% file: sections/30_prelim.tex
\section{Preliminary}

\paragraph{LRM Reasoning.} In this work, we focus on the task of mathematical reasoning. Given a question $x$, the goal of an LRM policy is to generate a response $y$ that contains step-by-step reasoning to derive the correct answer.
We assume access to a verifier $\mathcal{R}(x, y)$ that evaluates the correctness of a solution $y$ given the question $x$. In practice, such a verifier is implemented by matching the ground-truth answer and the model-generated answer. 
The LRM is a policy $\pi_\theta$ parameterized with $\theta$ and generates a sequence of reasoning tokens in an auto-regressive manner. 
Given a question distribution $\mathcal D$, the objective of the LRM is to maximize the probability of producing correct responses,
\begin{align}
J(\mathcal D, \theta)=\mathbb E_{x\sim \mathcal D,y\sim\pi_\theta(\cdot|x)}[\mathcal R(x, y)]
\end{align}

where $\theta$ is in a parameter space $\Theta$ and the response length $|y|$ is limited to the maximum length $L_{\text{max}}$. In practice, $\theta$ is usually obtained through applying fine-tuning approaches such as RL on a base LRM. Therefore we assume the existence of a base LRM $\theta_{\text{base}}$ and $\Theta$ to be the set of all LRMs that could be obtained by fine-tuning $\theta_{\text{base}}$ with any algorithm.

%% file: sections/40_length_scaling.tex
\section{Understanding the Limits of Efficient Reasoning}

\subsection{Defining Optimality in Token-Bounded Reasoning}
\label{sec:define-optimal}

\paragraph{Evaluating Reasoning under Token Budgets.} 
To assess optimal reasoning efficiency, we must evaluate the performance of an LRM $\pi_\theta$ under a fixed token budget $L$. Simply truncating an output after $L$ tokens, however, may lead to incomplete responses with no answers. To address this, we define a fallback mechanism following prior works~\citep{muennighoff2025s1simpletesttimescaling,fu2024efficiently}. If the reasoning trace $y \sim \pi_\theta(\cdot|x)$ exceeds $L$ tokens, the model is prompted to produce a final answer $a$ directly from the truncated trace $y_{:L}$,
\begin{align*}
a=\mathrm{Answer}(\pi_\theta,x,y)=\begin{cases}\pi_\theta(\cdot|x, y_{:L},[\text{The Final Answer is}])\quad&\text{if}\;|y|>L \\\mathrm{ExtractAnswer}(x,y)&\text{otherwise}\end{cases} 
\label{eq:answer-forcing}
\end{align*}

where $y_{:L}$ denotes the first $L$ tokens of the reasoning trace and $\mathrm{ExtractAnswer}(x,y)$ extracts the final answer from a complete trace.
 This approach allows consistent evaluation across different budgets, though it introduces a minor additional token cost in the truncation case. \gjx{For conciseness, we use $r(x, y_{:L};\theta)$ to represent the expected reward for $\pi_\theta$ on the response $y_{:L}$, i.e.}
 \begin{align}
 r(x,y_{:L};\theta)=\mathcal R(x, \mathrm{Answer}(\pi_\theta, x, y_{:L}))
 \end{align}

\paragraph{Length-Constrained Reward and Optimality.} We define the \emph{length-constrained reward} for a model $\pi_\theta$ over a question distribution $\mathcal D$ as the expected reward obtained when the model is restricted to a token budget $L$,
\begin{align}
J(\mathcal D,\theta, L)=\mathbb E_{x\sim \mathcal D}[\mathbb E_{y\sim\pi_\theta(\cdot|x)}[r(x,y_{:L};\theta)]]
\label{eq:length-constrained-reward}
\end{align}

The \emph{length-constrained optimal reward} then captures the best possible reward achievable by any model in a parameter space $\Theta$ under the same budget,
\begin{align}
J_{\text{optimal}}(\mathcal D, \Theta, L) = \max_{\theta\in\Theta}J(\mathcal D,\theta, L)
\end{align}

\subsection{Empirical Estimation of Reasoning Frontiers}
\label{sec:estimating-optimal}

\begin{figure*}[t]
\vspace{-8mm}
\centering     
    \centering
    \includegraphics[width=\linewidth]{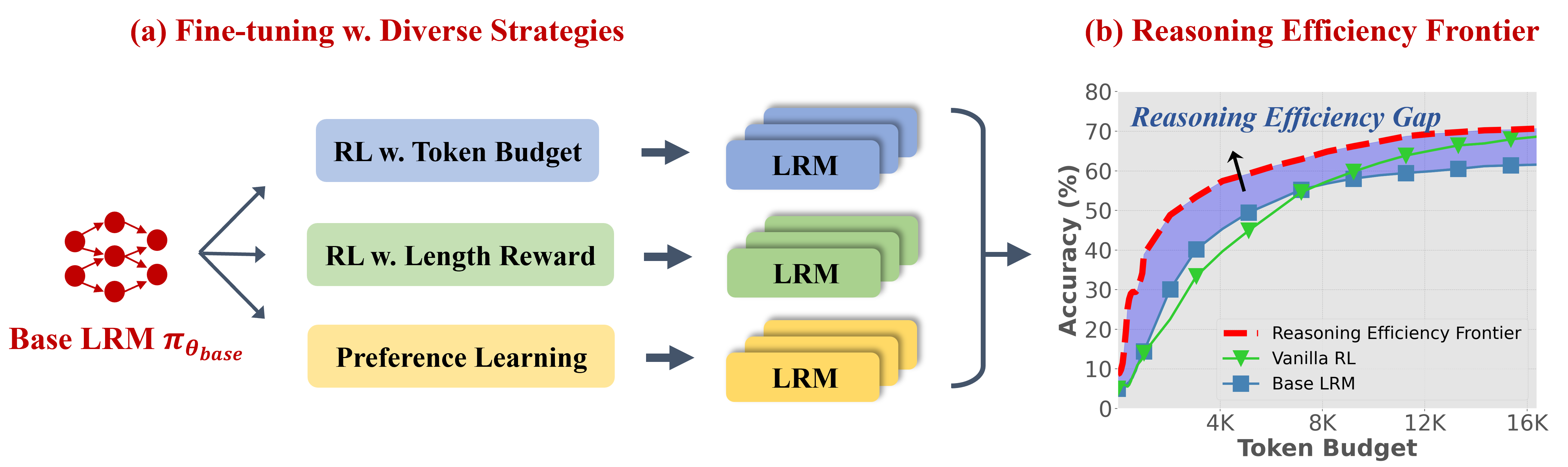}
    \vspace{-3mm}
    \caption{\textbf{Reasoning Efficiency Frontiers \& Reasoning Efficiency Gap.} (a) Starting from a base LRM $\pi_{\theta_{\text{base}}}$, we apply diverse fine-tuning strategies to obtain a large amount of LRMs. (b) We then compute the best achievable accuracy across varying token budgets to obtain the reasoning efficiency frontiers (Eq.~\ref{eq:full-optimal-reasoning}). \gjx{Reasoning Efficiency Gap (REG) is a unified metric that captures both accuracy and length by measuring the area between the budget-accuracy curve of an LRM and the efficiency frontiers.}}
    \label{fig:frontier-estimation}
    \vspace{-4mm}
\end{figure*}

We aim to characterize the optimal reasoning efficiency, denoted by $J_{\text{optimal}}(\mathcal{D}, \Theta, L)$, that reflects the best achievable reward at any token budget $L$ across all possible model parameters $\theta \in \Theta$. However, computing this optimal frontier exactly is infeasible in practice, as it requires exhaustively exploring all algorithms and training configurations. Instead, we construct an empirical reasoning efficiency frontier by fine-tuning a diverse set of models using existing approaches. Let $\hat{\Theta} = \{\theta_1, \ldots, \theta_m\} \subseteq \Theta$ denote the collection of parameters from $m$ fine-tuned models. Based on these, we define the empirical reasoning efficiency frontier as,

\begin{mydefinition}{Reasoning Efficiency Frontier}{reasoning-efficiency-frontier}
Given a parameter space $\Theta$, a set of model parameters $\hat\Theta=\{\theta_1,\cdots,\theta_m\}$ and a question distribution $\mathcal D$, we define the reasoning efficiency frontier $\hat J_{\text{optimal}}(\mathcal D,\hat \Theta, L)$ as
\begin{align}
\hat J_{\text{optimal}}(\mathcal D,\hat\Theta, L)=\max_{\theta\in\{\theta_1,\cdots,\theta_m\}}J(\mathcal D_t,\theta, L)\;\;\;\;\forall L\in[1,L_{\text{max}}]
\label{eq:full-optimal-reasoning}
\end{align}
\end{mydefinition}

Note that $\hat J_{\text{optimal}}(\mathcal D, \hat\Theta, L)$ serves as a lower bound of the optimal frontier since $\hat\Theta$ is a subset of $\Theta$,
$$\hat J_{\text{optimal}}(\mathcal D,\hat\Theta, L) \le J_{\text{optimal}}(\mathcal D,\Theta, L)$$

\paragraph{Diverse Fine-tuning Approaches.} To obtain a close approximation to the optimal frontier in Eq.~\ref{eq:full-optimal-reasoning}, we fine-tune models using a wide range of training strategies,

\begin{itemize}[topsep=5pt, leftmargin=7mm]
\vspace{-1em}
\item \textbf{Online RL with Token Budgets:} We conduct online RL training with different token budgets ranging from 512 to 32k. Fine-tuning an LRM with a token budget effectively enforces the LRM to reason with limited tokens~\citep{xu2025scalablechainthoughtselastic,hou2025thinkprune}.
\item \textbf{Online RL with Length Rewards.} We test various reward designs that promote concise yet accurate reasoning, including length-harmonizing rewards~\citep{luo2025o1} and length-group normalized rewards~\citep{arora2025training}.
\item \textbf{Preference Learning.} We apply SimPO~\citep{meng2024simpo} with preference datasets constructed via methods such as TOPS~\citep{yang2025thinkingoptimalscalingtesttimecompute} and DAST~\citep{shen2025dast}. For example, we contrast short correct responses with longer ones to promote conciseness~\citep{munkhbat2025self}.
\end{itemize}

\paragraph{Experiment Setup.} \gjx{We conduct our study using two classes of base LRMs, \emph{DeepSeek-R1-Distill-Qwen-1.5B/7B}~\citep{guo2025deepseek} and \emph{Qwen3-4B/8B}~\citep{yang2025qwen3technicalreport}.} Specially, for DeepSeek-R1-Distill-Qwen-1.5B/7B, we start from RL-fine-tuned versions and further fine-tune them using the strategies above. The evaluation is carried out on \gjx{a set of challenging mathematical reasoning benchmarks.} 
More training \gjx{and evaluation} details can be found in Sec.~\ref{sec:exp} and Appendix.~\ref{app:impl}.

\begin{figure*}[t]
\vspace{-2mm}
\centering   
    \centering
    \subfigure[DeepSeek-R1-Distill-Qwen-1.5B]{
    \includegraphics[width=0.43\linewidth]{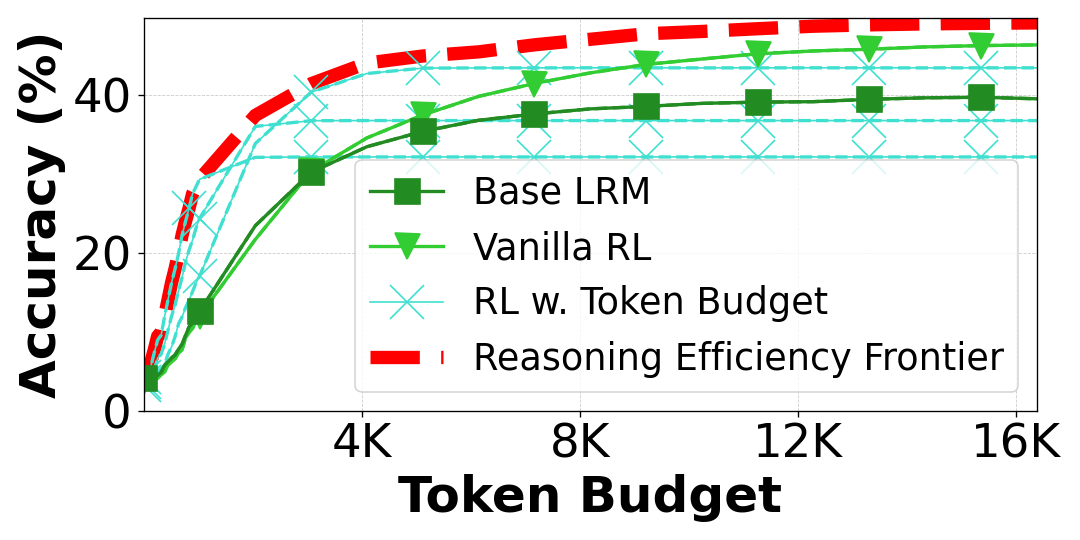}
    }
    \hspace{5mm}
    \subfigure[DeepSeek-R1-Distill-Qwen-7B]{
    \includegraphics[width=0.43\linewidth]{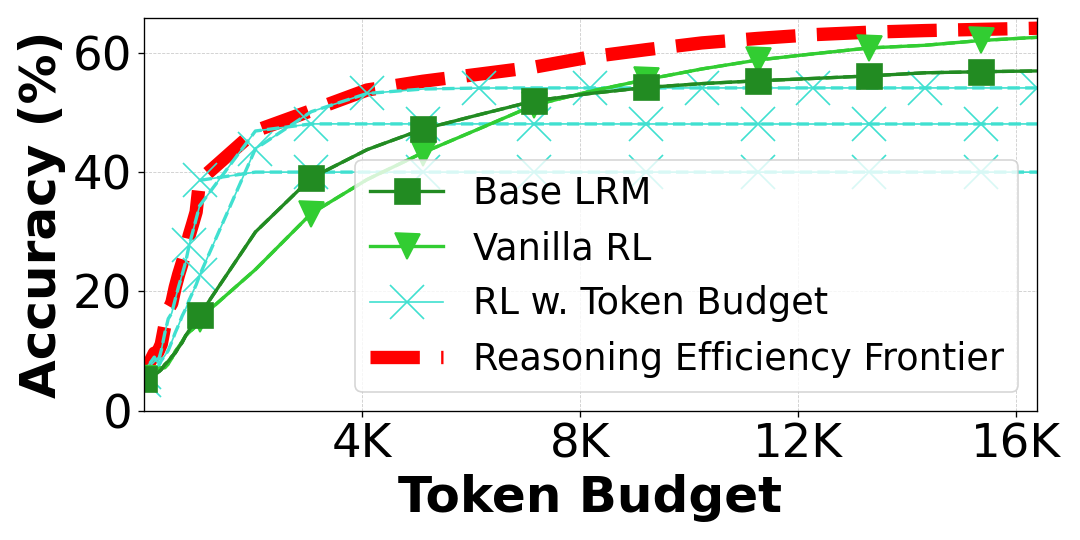}
    
    }
    \subfigure[Qwen3-4B]{
    \includegraphics[width=0.43\linewidth]{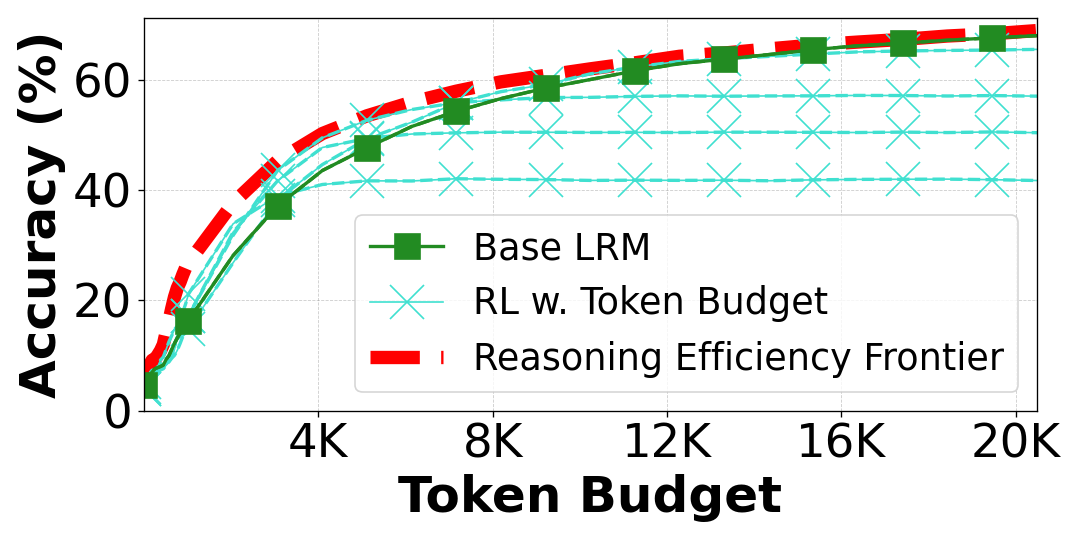}
    }
    \hspace{5mm}
    \subfigure[Qwen3-8B]{
    \includegraphics[width=0.43\linewidth]{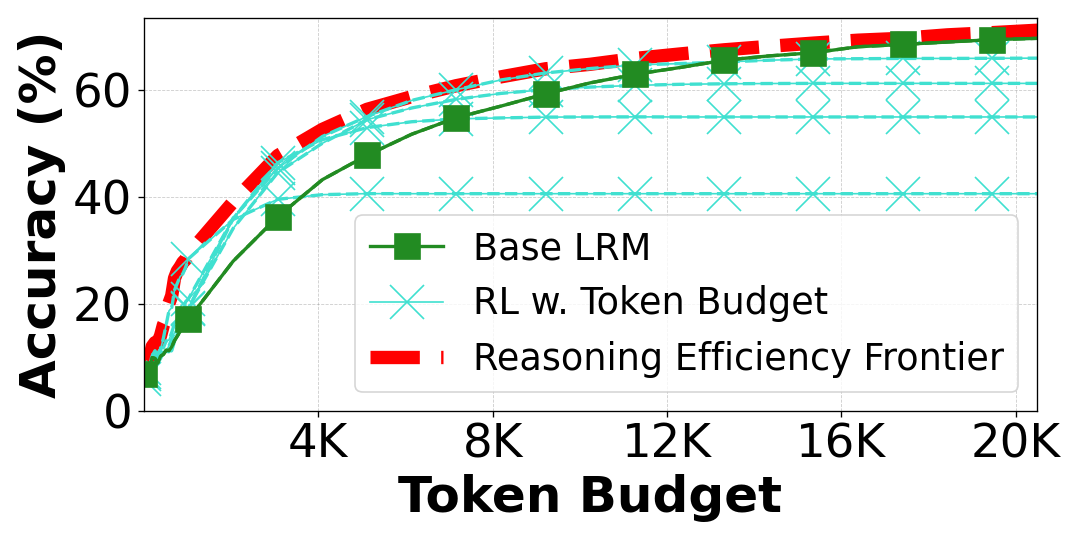}
    
    }
    \caption{\gjx{Reasoning efficiency frontiers for DeepSeek-R1-Distill-Qwen-1.5B/7B and Qwen3-4B/8B. Comparing the two classes of models, Qwen3 models have higher accuracy and are closer to the efficiency frontiers than DeepSeek-R1-Distill-Qwen models.}}
    \vspace{-5mm}
    \label{fig:frontiers}
\end{figure*}

\paragraph{Reasoning Efficiency Frontiers.} Based on the fine-tuned models, we are able to construct the reasoning efficiency frontiers for \gjx{ the base LRMs, as illustrated in Fig.~\ref{fig:frontiers}. Interestingly, we find that DeepSeek-R1-Distill-Qwen models are both far from the efficiency frontiers, with clear efficiency gaps. By contrast, Qwen3 models reach a similar accuracy as the efficiency frontiers and are much closer to the efficiency frontiers, allowing limited improvement space for reasoning efficiency.}

\paragraph{Measuring the Gap to Optimality.} Given the estimated reasoning efficiency frontiers $\hat J_{\text{optimal}}(\mathcal D,\Theta, L)$, we introduce the metric of \textbf{\emph{Reasoning Efficiency Gap  (REG)}} to quantify the distance of any LRM from reaching the optimal reasoning efficiency.

\begin{mydefinition}{Reasoning Efficiency Gap  (REG)}{reg}
Given any LRM $\pi_{\theta}$ and the estimated reasoning efficiency frontier $\hat J_{\text{optimal}}(\mathcal D,\hat\Theta, L)$, we define Reasoning Efficiency Gap as,
\begin{align}d_{\text{REG}}(\theta, \mathcal D,\hat\Theta) = \sum_{L=1}^{L_{\text{max}}}\hat J_{\text{optimal}}(\mathcal D, \hat\Theta, L) - J(\mathcal D,\theta, L)
\label{eq:reg}
\end{align}
\end{mydefinition}

\gjx{Note that, when $\hat J_{\text{optimal}}(\mathcal D, \hat \Theta, L)$ and $J(\mathcal D,\theta, L)$ are both non-decreasing,  REG could equivalently represent the average amount of wasted tokens across all accuracies.}

\paragraph{Advantages of REG.} \gjx{As a single metric, REG unifies both the accuracy and length aspects. At the same time, REG quantitatively assesses the gap of reasoning efficiency between an LRM and efficiency frontiers. 
Most of the existing evaluation metrics only focus on one of the accuracy and length aspects. For example, \citet{pu2025thoughtterminatorbenchmarkingcalibratingmitigating} proposes to evaluate reasoning redundancy by computing the length difference between the longest and the shortest response, without considering correctness of the generated responses. We also note that \citet{chen2024not} introduces an outcome efficiency metric that measures the ratio between the position of first reaching the correct answer and the the overall length for correct responses. Although this outcome efficiency metric takes both accuracy and length into consideration, it has the risk of being hacked if the LRM only produces an answer at the end of the response.}

\section{Methodology}

\subsection{Boosting Reasoning Efficiency by Optimizing Length-Constrained Rewards}

\paragraph{Optimizing Length-Constrained Rewards.}
To minimize the reasoning efficiency gap, a straightforward idea is to optimize the length-constrained rewards under all token budgets to enhance the reasoning efficiency of $\pi_{\theta_{\tt{base}}}$, leading to the efficiency objective,
\begin{align}
    \mathscr{L}_{\text{Efficiency}}(\theta,\mathcal D)=\sum_{L=1}^{L_{\tt{max}}}J(\mathcal D, \theta, L)
\label{eq:optimal-max}
\end{align}
where $L_{\tt{max}}$ is the maximum generation length.
However, directly optimizing Eq.~\ref{eq:optimal-max} is computationally impractical. Evaluating $J(\mathcal D, \theta, L)$ for each budget $L \in [1, L_{\tt{max}}]$ requires separate inference runs to evaluate truncated responses as discussed in Sec.~\ref{sec:define-optimal}. As a result, each training example would require up to $L_{\tt{max}}$ additional LRM generations, leading to increased compute time and memory usage. 

\subsection{\methodname}

\paragraph{{\method}.} We introduce \emph{{\methodname} ({\method})}, a class of an efficient training algorithms that optimize an approximation for the objective in Eq.~\ref{eq:optimal-max}. In {\method}, instead of optimizing the length-constrained reward in all token budgets, we approximate the objective with a small set of selected token budgets $L_1, \cdots, L_{N}$ to ensure high training efficiency. 
 Specifically, following the Trapezoidal rule in numerical integration, we could approximate the objective in Eq.~\ref{eq:optimal-max} with,
\begin{align}
\sum_{L=1}^{L_{\tt{max}}}J(\mathcal D, \theta, L) &\approx 
\sum_{i=1}^N\frac{L_{i+1}-L_{i-1}}2J(\mathcal D,\theta, L_i)+\frac{L_1}{2}\cdot J(\mathcal D,\theta, 0) + \frac{L_{\text{max}}-L_N}{2}\cdot J(\mathcal D,\theta, L_{\text{max}})\\
&=f(\mathcal D,\theta, \{L_1,\cdots, L_N\})
\label{eq:approx-optimal-max}
\end{align}

where we assume $L_0=0$ and $L_{N+1}=L_{\text{max}}$ and $f(\mathcal D,\theta, \{L_1,\cdots, L_N\})$ denotes the approximated objective with token budgets $L_1,\cdots,L_N$. As the number of selected token budgets $N$ increases, $f(\mathcal D,\theta, \{L_1,\cdots, L_N\})$ would become closer and closer to the objective in Eq.~\ref{eq:optimal-max}. 

The approximated objective $f(\mathcal D,\theta, \{L_1,\cdots, L_N\})$ could be equivalently represented as an RL objective with dense rewards, leading to the objective of {\method},
\begin{align}
\textbf{REO-RL:}&&\mathscr{L}_{{\method}}(\theta,\mathcal D)= \mathbb E_{x\sim \mathcal D}\left[\mathbb E_{y\sim \pi_\theta(\cdot|x)}\left[\sum_{i=1}^{N+1}c_i\cdot r(x,y_{:L_i};\theta)\right]\right]
\label{eq:reo-rl}
\end{align}
where $c_i=\frac{L_{i+1}-L_{i-1}}{2}$ for $1\le i\le N$ 
and $c_{N+1}=\frac{L_{\text{max}}-L_N}{2}$ are the coefficient for the $i$-th token budget following Eq.~\ref{eq:approx-optimal-max}. 

\begin{figure*}[t]
\vspace{-2mm}
\centering 
    \centering
    \subfigure[]{
    \includegraphics[width=0.4\linewidth]{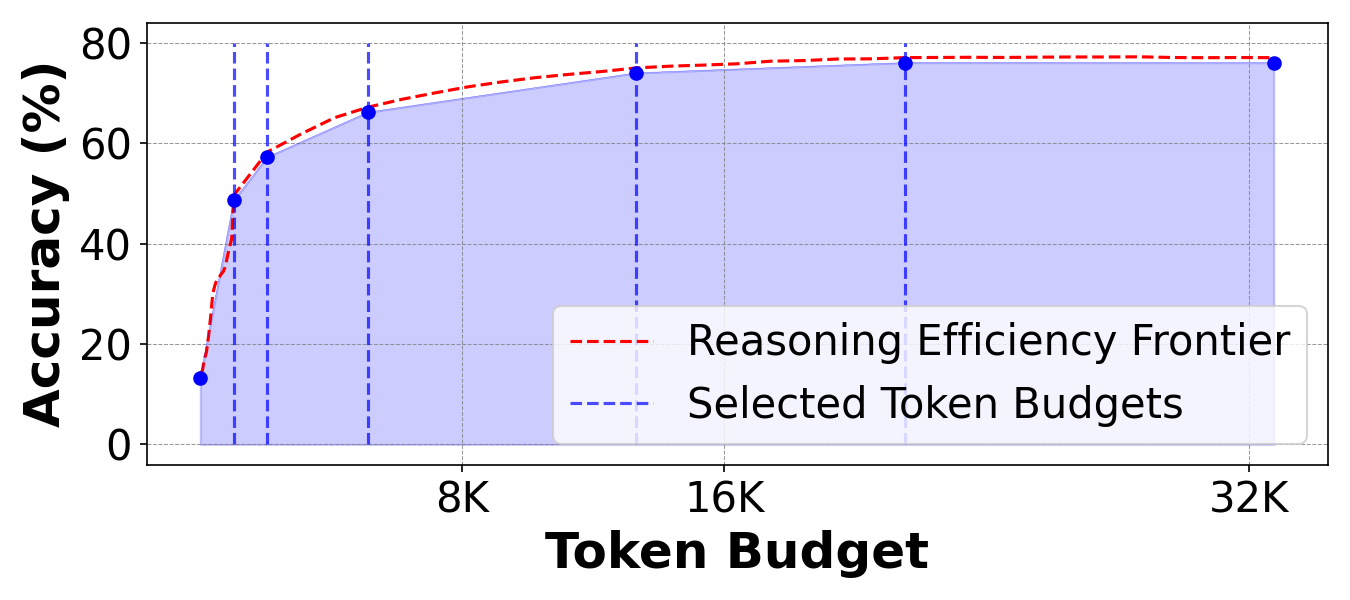}
    \label{fig:greedy-selection}
    }
    \hspace{2mm}
    \subfigure[]{
    \includegraphics[width=0.4\linewidth]{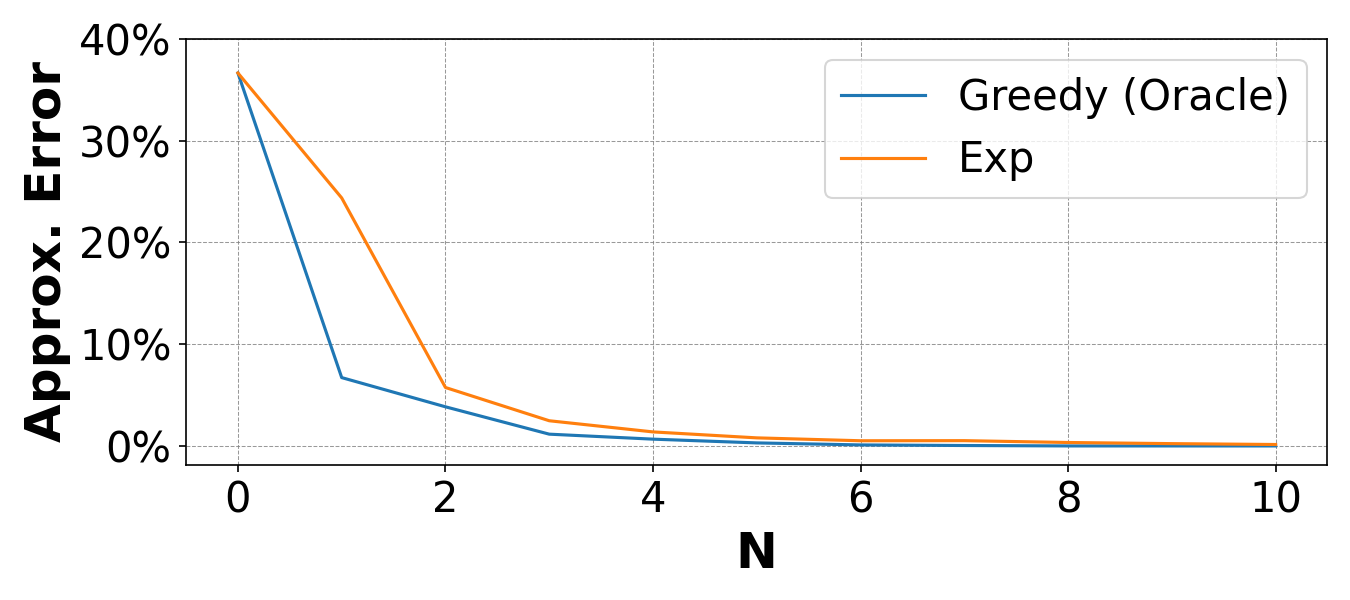}
    \label{fig:N-vs-approx}
    }
    \caption{Selection of Token Budgets in {\method}. (a) Token budgets selected roughly follow an exponential pattern. (b) Both the oracle greedy approach and the exponential approach achieve lower approximation errors with a few token budgets. }
    \vspace{-5mm}
\end{figure*}

Different sets of token budgets $L_1,\cdots,L_N$ would cause different approximation error in Eq.~\ref{eq:approx-optimal-max}. Ideally, the approximation error induced by the selected token budgets should be low to ensure that {\method} aligns with the original objective in Eq.~\ref{eq:full-optimal-reasoning}. 

\paragraph{{\method} (Oracle).}  Note that the original objective in Eq.~\ref{eq:full-optimal-reasoning} is bounded by the theoretical reasoning efficiency frontier in Sec.~\ref{sec:estimating-optimal}, i.e. $\sum_{L=1}^{L_{\text{max}}}J(\mathcal D,\theta, L)\le \sum_{L=1}^{L_{\text{max}}}J_{\text{optimal}}(\mathcal D,\Theta, L)$. A natural idea is to determine the optimal token budget selection scheme based on the estimated reasoning efficiency frontiers. Specifically, for any set of token budgets $L_1,L_2,\cdots, L_N$, $\sum_{L=1}^{L_{\text{max}}}\hat J_{\text{optimal}}(\mathcal D,\hat\Theta, L)$ could be approximated in a similar way as Eq.~\ref{eq:approx-optimal-max},
\begin{align}
f_{\text{optimal}}(\mathcal D, \hat\Theta, \{L_1,\cdots, L_N\})&= \sum_{i=1}^N \frac{L_{i+1}-L_{i-1}}{2}\hat J_{\text{optimal}}(\mathcal D,\hat\Theta, L_i)+\frac{L_1}{2}\cdot \hat J_{\text{optimal}}(\mathcal D,\hat\Theta, 0)\\\;\; &+ \frac{L_{\text{max}}-L_N}{2}\cdot \hat J_{\text{optimal}}(\mathcal D,\hat\Theta, L_{\text{max}})
\label{eq:optimal-approx}
\end{align}
where $f_{\text{optimal}}(\mathcal D,\hat\Theta, \{L_1,\cdots,L_N\})$ denotes the approximated value for $\sum_{L=1}^{L_{\text{max}}}\hat J_{\text{optimal}}(\mathcal D,\hat\Theta, L)$ given token budgets $L_1,\cdots,L_N$.

We adopt a greedy approach that iteratively selects the token budget with the lowest approximation error. Note that this is an oracle approach since the reasoning efficiency frontiers should be known in advance. This greedy selection approach leads to an oracle algorithm,  \emph{\textbf{{\method} (Oracle)}}. In {\method} (Oracle), the token budgets $L_1,\cdots, L_N$ are selected according to,
\begin{align*}
L_i=\arg\min_{L'}\mid f_{\text{optimal}}(\mathcal D,\hat\Theta, \{L_1, \cdots, L_{i-1},L'\}) - \sum_{L=1}^{L_{\text{max}}}\hat J_{\text{optimal}}(\mathcal D,\hat\Theta,L)\mid
\end{align*}
The oracle greedy approach could produce a set of token budgets that achieves low approximation error in Fig.~\ref{fig:greedy-selection}. As illustrated in Fig.~\ref{fig:N-vs-approx}, the approximation error gradually degrades with more token budgets. Notably,  the approximation error could be lower than $1\%$ with $N\ge 5$.

\paragraph{{\method} (Exp).} However, in cases when the reasoning efficiency frontiers are unknown, it is infeasible to apply {{\method} (Oracle)}. We observe that token budgets selected by the greedy selection approach roughly follow an exponentially spaced pattern as shown in Fig.~\ref{fig:greedy-selection}. Therefore, we propose to adopt an exponentially spaced scheme for token budget selection, leading to the algorithm, \textbf{\emph{{\method} (Exp)}}, that selects a set of exponentially spaced token budgets,
\begin{align*}
L_i=L_{\text{min}}\cdot \left({L_{\text{max}}}/{L_{\text{min}}}\right)^{\frac{i-1}{N}}
\end{align*}
where $L_{\text{min}}/L_{\text{max}}$ are the minimum/maximum token budgets.

%% file: sections/60_experiments.tex
\section{Experiments}

\subsection{Experimental Setup}
\label{sec:exp}

\paragraph{Models, Datasets \& Metrics.} We use \gjx{DeepSeek-R1-Distill-Qwen-1.5B\&7B and Qwen3-4B\&8B} as the base LRMs. For training, we adopt a mixture of training data consisting of 135k problems sourced from DeepScaleR~\cite{deepscaler2025} and AReaL~\cite{areal2025} For evaluation, we use \gjx{several challenging mathematical benchmarks: AMC 2023, AIME 2024 \& 2025, and Minerva Math~\citep{lewkowycz2022solvingquantitativereasoningproblems}. }
We report the average accuracy of 32 responses generated with temperature $T=0.6$ and top\_p$=0.95$ with maximum length $L_{\max}=32K$. The main results are averaged over three benchmarks. When evaluating REG, we use a smaller length $L_{\text{max}}=16K$.

\paragraph{{\method} \& Baselines.} For {\method}, besides {\method} (Exp) and {\method} (Oracle), we further consider {\method} (Q-Spec), a variant of {\method} that optimizes rewards in question-specific minimum reasoning token budgets\footnote{For more details on {\method} (Q-Spec), please refer to Appendix~\ref{app:method}}.

\begin{itemize}[topsep=5pt, leftmargin=7mm]
\item\textbf{Online RL with Length-based Rewards}: We consider online RL with length-based rewards, including length-harmonizing rewards~\citep{luo2025o1}, and length group normalized rewards~\citep{arora2025training}. We make comparison with Meta Reinforcement Fine-tuning (MRT)~\citep{qu2025optimizing}, which minimizes regret for individual steps to enhance reasoning efficiency. Additionally, we also adopt online RL with hard token budgets as a baseline, which is also adopted in recent works~\citep{hou2025thinkprune,xu2025scalablechainthoughtselastic}. \gjx{We note that a recent work BRPO~\citep{qi2025optimizinganytimereasoningbudget} optimizes rewards under linearly spaced token budgets, which is similar to a variant of {\method} studied in the ablation study (Sec.~\ref{sec:ablation}).}\footnote{A more detailed discussion between BRPO and {\method} is provided in Appendix.~\ref{app:discuss}}

\item \textbf{Hybrid Reasoning.} \gjx{We use HGPO~\citep{jiang2025thinkneedlargehybridreasoning} as a representative hybrid reasoning baseline that trains the LRM to adaptively select between the thinking and no-thinking modes. There are also recent works sharing the similar idea of encouraging no-thinking mode when no-thinking mode achieves competitive accuracy~\citep{fang2025thinklessllmlearnsthink,zhang2025adaptthinkreasoningmodelslearn,tu2025learningthinkshapingadaptive}.
}

\item\textbf{Supervised Fine-Tuning}: For each problem in the training data, we generate multiple responses and perform SFT on the shortest correct one. Two strategies are used for data generation. The first strategy is direct generation with the problems as inputs~\citep{munkhbat2025self}. The second strategy follows TOPS to prompt the LRM with different levels of reasoning efforts~\citep{yang2025thinkingoptimalscalingtesttimecompute}. 

\item\textbf{Preference Learning}: We apply SimPO~\citep{meng2024simpo} on various preference datasets. In $SimPO_{shortest}$, we use the shortest correct response and the longest response as the preference pairs. We also follow TOPS~\citep{yang2025thinkingoptimalscalingtesttimecompute} and DAST~\citep{shen2025dast} to construct preference datasets.
\end{itemize}

\vspace{-1em}
We have also tried RL-based length control methods such as ~\citep{aggarwal2025l1,xu2025scalablechainthoughtselastic} but find these methods only achieve successful length control under low token budgets and yield similar length-accuracy trade-off as RL w. Token Budgets. More details about the baselines and our other investigations could be found in Appendix.~\ref{app:baseline}.

\begin{figure*}[t]
\vspace{-5mm}
\centering     
    \centering
    \subfigure[DeepSeek-R1-Distill-Qwen-1.5B]{
    \includegraphics[width=0.42\linewidth]{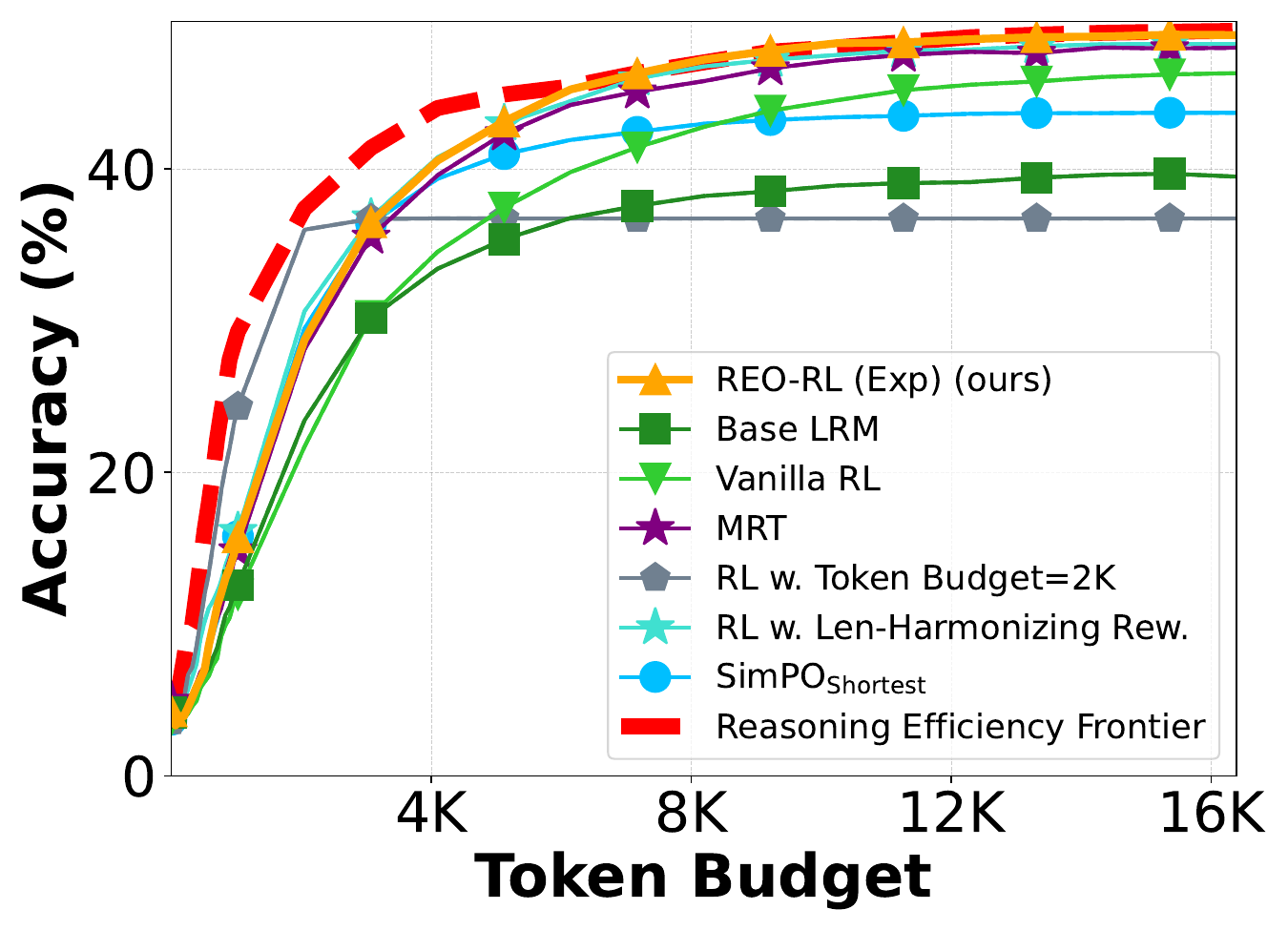}
    }
    \hspace{-3mm}
    \subfigure[DeepSeek-R1-Distill-Qwen-7B]{
    \includegraphics[width=0.42\linewidth]{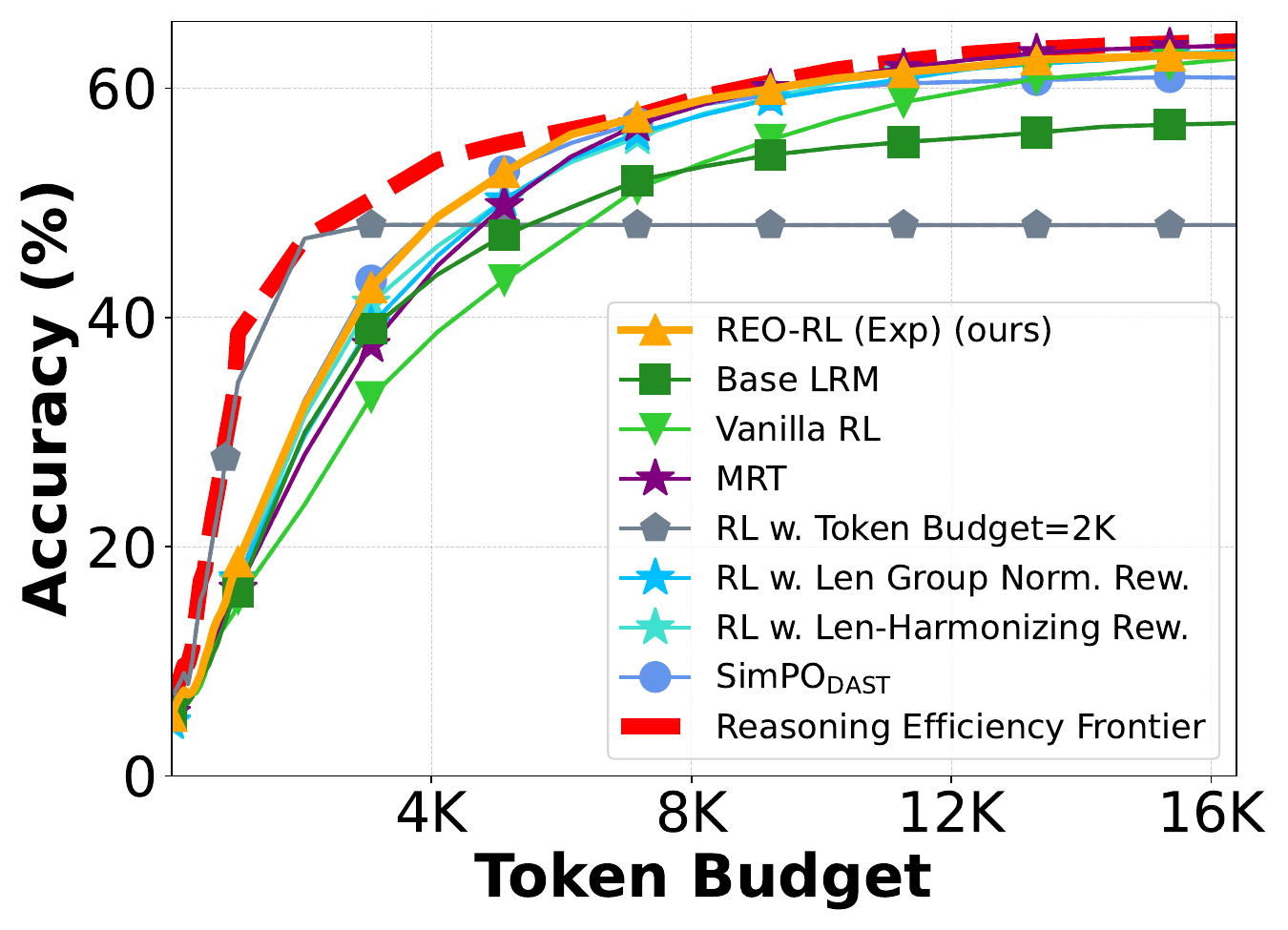}
    }
    \subfigure[Qwen3-4B]{
    \includegraphics[width=0.42\linewidth]{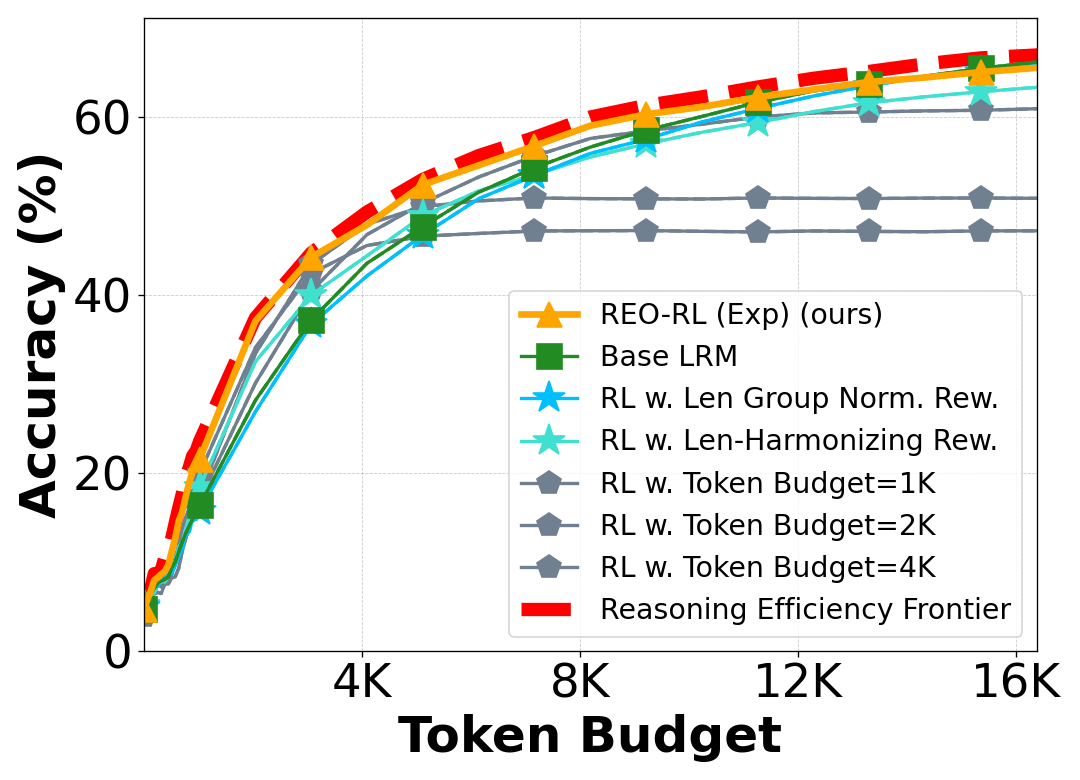}
    }
    \hspace{-3mm}
    \subfigure[Qwen3-8B]{
    \includegraphics[width=0.42\linewidth]{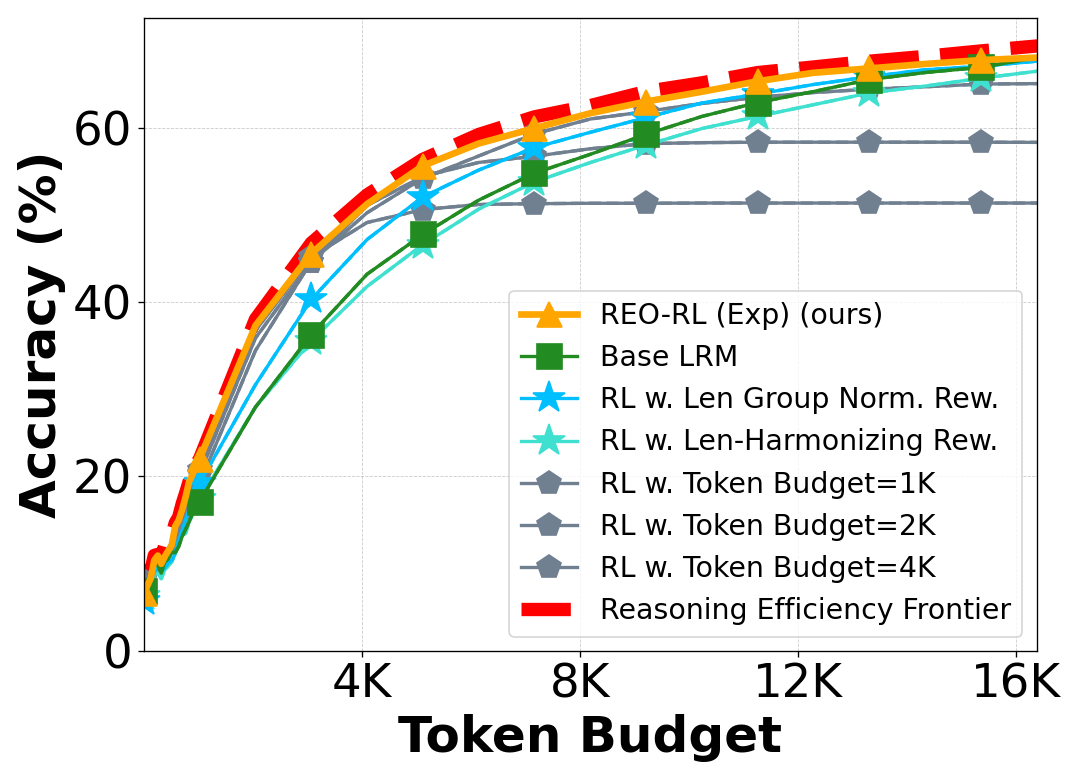}
    }
    \vspace{-4mm}
    \caption{Performance comparison of {\method} (Exp) with representative baseline methods \gjx{under 16K token budget}. {\method} (Exp) outperforms baselines in terms of reasoning efficiency. For DeepSeek-R1-Distill-Qwen models, there still exist large performance gaps under tight token constraints. For Qwen3 models, though {\method} matches the efficiency frontiers under the 16K token budget, accuracy loss still exists (See Tab.~\ref{tab:main-result}). 
    Results are averaged over AMC 2023, AIME 2024 \& 2025, and Minerva Math.}
    \label{fig:main-result}
\end{figure*}

\begin{figure*}[h]
\centering
    \includegraphics[width=0.5\linewidth]{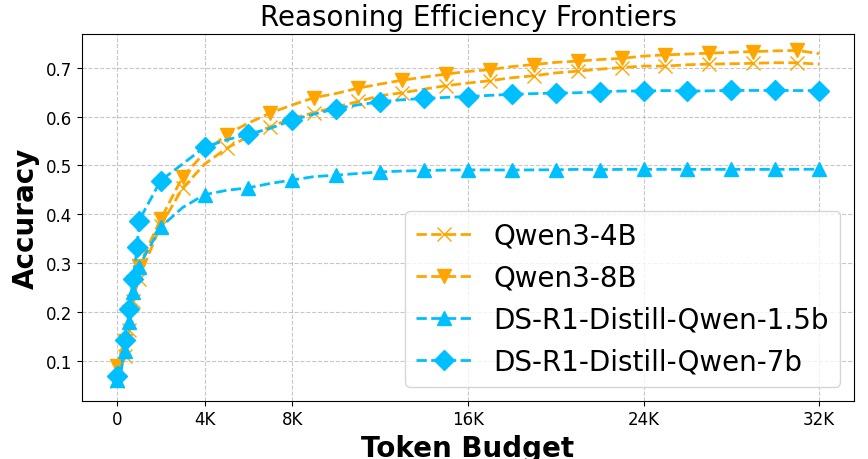}
    \caption{\gjx{Reasoning efficiency frontiers of DeepSeek-R1-Distill-Qwen-1.5B/7B and Qwen3-4B/8B. Qwen3 models obtain higher accuracy. Interestingly, DeepSeek-R1-Distill-Qwen-7B could reason more efficiently than Qwen3 models under the 4K token budget.}}
    \label{fig:all-frontiers}
\end{figure*}

\paragraph{Training Details.} For {\method} and all baselines, we implement all methods based on the AReaL framework~\citep{areal2025}. For DeepSeek-R1-Distill-Qwen-1.5B/7B, we fine-tune the corresponding RL-trained versions. Specifically, we use AReaL-Boba-RL-1.5B~\citep{areal2025} and SkyWork-OR1-Math-7B~\citep{skywork-or1-2025} as the starting points for further fine-tuning in 1.5B and 7B experiments, respectively. \gjx{For Qwen3 models, we directly apply RL training on the base LRMs.} For {\method} (Exp) and {\method} (Oracle), we use $N=5$ token budgets since $N=5$ already obtains sufficiently accurate approximations as illustrated in Fig.~\ref{fig:N-vs-approx}. For more training details, please refer to Appendix.~\ref{app:impl}. 

\subsection{Main Results}

\begin{figure*}[t]
\vspace{-5mm}
\centering
    \includegraphics[width=0.99\linewidth]{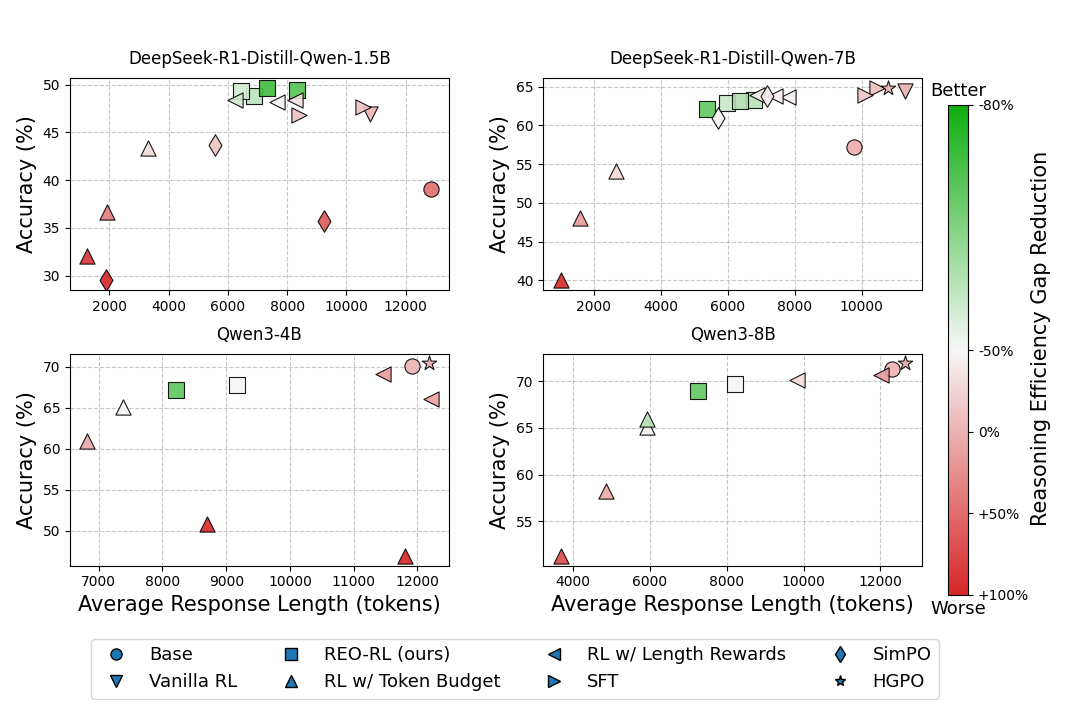}
    \caption{\gjx{\textbf{REG effectively captures the trade-off between accuracy and response length.} Achieving a low REG require both competitive accuracy and short response length. By minimizing the efficiency gap, {\method} outperforms baselines in terms of reasoning efficiency.}}
    \label{fig:acc-len-reg}
\vspace{-3mm}
\end{figure*}

Fig.~\ref{fig:main-result} presents the budget-accuracy curves of LRMs fine-tuned using {\method} (Exp) and a range of representative baseline methods.\footnote{For clearer visualization, we include only a set of representative approaches and plot their performance under limited token budgets.} 
\gjx{Fig.~\ref{fig:acc-len-reg} further illustrates the accuracy, response length, and REG reduction for different classes of methods. }
Full quantitative results are detailed in Tab.~\ref{tab:main-result}. We highlight several key conclusions,

\textbf{There exist fundamental gaps between existing approaches and the frontiers.} As shown in Fig.~\ref{fig:main-result}, several baselines can approach or even reach the reasoning efficiency frontier given sufficient token budgets. However, they could require significantly more tokens than the efficiency frontier to achieve moderate-level accuracies. 
On the other hand, some methods, such as RL with Token Budget, perform well under tight token constraints but exhibit substantially lower overall accuracy. \gjx{Notably, our proposed approach, {\method}, could match the efficiency frontiers under low token budget for Qwen3 models, but still with slight drop in the overall accuracy, as illustrated in Tab.~\ref{tab:main-result}.} Our benchmarking results suggest that \emph{optimizing the base LRM to precisely match reasoning efficiency frontiers remains an open problem}.

\textbf{REG effectively captures the trade-off between accuracy and response length.} As shown in Fig.~\ref{fig:acc-len-reg}, achieving low REG requires balancing high accuracy with concise output. For example, {\method} maintains strong accuracy while generating compact responses, resulting in significant reduction in REG. In contrast, methods that optimize only one aspect, such as minimizing length at the cost of accuracy, fail to achieve low REG. RL with Token Budgets, despite producing shorter responses, suffers from significant accuracy loss.

\textbf{Finally, {\method} consistently outperforms baselines in terms of reasoning efficiency.} Compared to baselines, the of {\method} are much closer to the reasoning efficiency frontiers, exhibiting much more reduction in the REG. {\method} maintains high accuracy while generating shorter outputs. 
\gjx{{\method} (Exp) consistently reduces the REG by at least 50\% across all base LRMs. For Qwen3 models, baselines that apply RL with length-based rewards could not consistently reduce the efficiency gap. In fact, we find that RL w. Length Harmonizing Rewards even enlarges the efficiency gap for Qwen3-4B/8B models. }

\input{tables/main_results_v3}

\gjx{Besides these major findings, we also have the following observations,}

\textbf{Vanilla RL does not yield consistent improvements across budgets.} Although Vanilla RL improves overall accuracy relative to the base LRM, it fails to provide stable gains across different token budgets. In the 7B experiments, its accuracy is lower than the base LRM when the token budget is below 8K.

\textbf{HGPO does not necessarily improve reasoning efficiency with adaptive thinking mode switch.} \gjx{We find HGPO could enhance the overall accuracy but not necessarily improves the reasoning efficiency and reduces response length. In our experiments, we observe that the LRM would learn to achieve higher rewards in the no-thinking mode, along with an increasing response length during the training process of HGPO. This indicates the LRM reverts back to the thinking mode even with a no-thinking prompt, i.e. performing long-CoT reasoning outside of the ``<think>...</think>'' tags.}

\subsection{Comparison with Frontier LRMs}
\label{sec:frontier_LRM}

\input{tables/frontier}

To further evaluate the reasoning efficiency of LRMs trained with {\method} (Exp) in comparison to advanced LRMs, we conduct a controlled analysis focused on the correct response length. Since frontier LRMs and those fine-tuned with {\method} (Exp) differ significantly in overall accuracy, we construct a balanced subset of 71 questions from the test set, where all models achieve an accuracy exceeding 50\%. For each model, we compute the average length of correct reasoning traces for each problem. The final length metric for each model is then obtained by averaging the correct response lengths across all 71 questions. As shown in Tab.~\ref{tab:frontier}, the model obtained through fine-tuning DeepSeek-R1-Distill-Qwen-7B with {\method} (Exp) exhibits more concise reasoning patterns than frontier LRMs, with a much shorter response length.

\subsection{Ablation Study of {\method}}
\label{sec:ablation}

\input{tables/ablation}

We conduct an ablation study on the design choices of {\method} using the DeepSeek-R1-Distill-Qwen-7B model to better understand its flexibility and performance characteristics. We demonstrate that {\method} can maintain competitive performance across different configurations.

\textbf{Token budget selection strategy.} We evaluate {\method} (Oracle) that adopts the oracle greedy strategy. This variant achieves slightly better performance than {\method} (Exp), reflected in a lower REG. We also investigate linearly spaced token budgets, which leads to less optimal efficiency and higher REG.

\textbf{Coefficient $c_i$ in {\method} objective.} Setting all coefficients $c_i$ uniformly to 1 leads the model to align more closely with the efficiency frontier and shorter response lengths. However, this approach comes at the cost of reduced overall accuracy.

\textbf{Number of selected token budgets $N$.} We increase the number of selected token budgets to $N = 10$ to explore whether a finer granularity improves performance. In practice, we observe that using more token budgets results in slower convergence and a higher rate of training instability. Consequently, this configuration produces weaker results overall.

\textbf{Question-Specific Oracle Budgets.} We investigate a question-specific oracle strategy that assigns two token budgets to each problem, the minimum reasoning token budget derived through the frontier estimation experiments (Sec.~\ref{sec:estimating-optimal}) and the full budget $L_{\text{max}}$. This oracle approach performs better than {\method} (Exp) on DeepSeek-R1-Distill-Qwen models but leads to less REG reduction than {\method} (Exp) on Qwen3 models.

%% file: tables/main_results_v3.tex
\begin{table}[h]
\centering
    \resizebox{0.99\textwidth}{!}{
\begin{tabular}{@{}c|ccc|ccc|ccc|ccc|ccc@{}}
\toprule
\multirow{2}{*}{Method} & \multicolumn{3}{c}{AIME 2024} & \multicolumn{3}{c}{AIME 2025} &\multicolumn{3}{c}{AMC 2023} & \multicolumn{3}{c}{Minerva Math} & \multicolumn{3}{c}{Average}\\
 & Acc (\%) $\uparrow$ & Len $\downarrow$ & REG $\downarrow$ & Acc (\%) $\uparrow$ & Len $\downarrow$ & REG $\downarrow$  & Acc (\%) $\uparrow$ & Len $\downarrow$ & REG $\downarrow$  & Acc (\%) $\uparrow$ & Len $\downarrow$ & REG $\downarrow$ & Acc (\%) $\uparrow$ & Len $\downarrow$ & REG $\downarrow$   \\
\midrule
\multicolumn{16}{c}{\emph{DeepSeek-R1-Distill-Qwen-1.5B}} \\
\midrule

Base LRM  &  29.2 & 16757.1 & 2239.0 & 23.5 & 16577.6 & 1485.9 & 71.6 & 9956.8 & 2425.1 & 32.0 & 8212.0 & 970.7 & 39.1 & 12875.9 & 1780.2\\
\midrule

Vanilla RL  &  42.2 & 12902.2 & 1266.2 & 28.2 & 13984.8 & 1077.7 & 81.9 & 8044.4 & 1742.6 & 35.2 & 8305.8 & 641.3 & 46.9 & 10809.3 & 1181.9\\
\midrule


\rowcolor{bg!70} REO-RL (Exp) (ours)  &  42.9\bonus{0.7} & 8780.3 & \ddrop{46.6\%} & 31.9\bonus{3.6} & 8630.5 & \ddrop{64.4\%} & 84.5\bonus{2.7} & 4990.8 & \ddrop{63.6\%} & 36.1\bonus{0.8} & 5094.7 & \ddrop{51.7\%} & 48.8\bonus{2.0} & 6874.1 & \ddrop{57.6\%}\\

\rowcolor{bg!70} REO-RL (Oracle) (ours) &  42.5\bonus{0.3} & 8443.8 & \ddrop{30.0\%} & 34.7\bonus{6.5} & 8288.7 & \ddrop{85.4\%} & 84.4\bonus{2.5} & 4626.6 & \ddrop{53.1\%} & 35.8\bonus{0.5} & 4427.8 & \ddrop{56.5\%} & 49.3\bonus{2.5} & 6446.7 & \ddrop{54.7\%}\\
\rowcolor{bg!70} REO-RL (Q-Spec) (ours)  &  46.6\bonus{4.4} & 9162.3 & \ddrop{90.7\%} & 32.7\bonus{4.5} & 8832.4 & \ddrop{81.4\%} & 84.1\bonus{2.2} & 5338.5 & \ddrop{66.6\%} & 35.4\bonus{0.2} & 5979.7 & \ddrop{45.0\%} & 49.7\bonus{2.8} & 7328.2 & \ddrop{73.5\%}\\

RL w. Token Budget=1K  &  17.3\drop{24.9} & 1487.4 & \bbonus{170.4\%} & 13.5\drop{14.7} & 1282.8 & \bbonus{140.7\%} & 65.6\drop{16.2} & 1111.3 & \bbonus{43.1\%} & 32.0\drop{3.2} & 1108.8 & \ddrop{24.1\%} & 32.1\drop{14.8} & 1247.6 & \bbonus{90.3\%}\\
RL w. Token Budget=2K  &  22.4\drop{19.8} & 2423.9 & \bbonus{114.2\%} & 17.5\drop{10.7} & 1966.0 & \bbonus{90.6\%} & 73.1\drop{8.8} & 1547.1 & \ddrop{13.6\%} & 33.9\drop{1.3} & 1671.8 & \ddrop{57.0\%} & 36.7\drop{10.1} & 1902.2 & \bbonus{38.5\%}\\
RL w. Token Budget=4K  &  31.4\drop{10.8} & 3991.0 & \bbonus{25.8\%} & 26.1\drop{2.1} & 3770.8 & \ddrop{17.3\%} & 80.9\drop{0.9} & 2561.2 & \ddrop{60.3\%} & 35.2$_{0.0}$ & 2942.8 & \ddrop{65.5\%} & 43.4\drop{3.5} & 3316.5 & \ddrop{28.2\%}\\

RL w. Len Group Norm. Rew.  &  44.4\bonus{2.2} & 10486.0 & \ddrop{37.5\%} & 30.7\bonus{2.5} & 10599.6 & \ddrop{33.7\%} & 82.7\bonus{0.9} & 5911.5 & \ddrop{30.7\%} & 35.8\bonus{0.6} & 6057.2 & \ddrop{34.6\%} & 48.4\bonus{1.5} & 8263.6 & \ddrop{33.3\%}\\
RL w. Len-Harmonizing Rew.  &  41.5\drop{0.7} & 9119.2 & \ddrop{31.3\%} & 31.6\bonus{3.3} & 8113.9 & \ddrop{60.0\%} & 85.1\bonus{3.2} & 3993.2 & \ddrop{65.1\%} & 35.4\bonus{0.2} & 3721.0 & \ddrop{69.6\%} & 48.4\bonus{1.5} & 6236.8 & \ddrop{55.5\%}\\
MRT  &  42.9\bonus{0.7} & 9797.8 & \ddrop{45.3\%} & 29.8\bonus{1.6} & 9823.0 & \ddrop{41.1\%} & 84.9\bonus{3.0} & 5607.8 & \ddrop{54.2\%} & 35.0\drop{0.2} & 5445.0 & \ddrop{26.1\%} & 48.2\bonus{1.3} & 7668.4 & \ddrop{45.0\%}\\

$\text{SFT}_\text{Shortest}$  &  43.2\bonus{1.0} & 12898.1 & \ddrop{11.1\%} & 29.9\bonus{1.7} & 14221.8 & \ddrop{22.3\%} & 82.5\bonus{0.6} & 7513.7 & \ddrop{9.6\%} & 35.2\drop{0.1} & 7641.9 & \ddrop{11.5\%} & 47.7\bonus{0.8} & 10568.9 & \ddrop{13.2\%}\\
$\text{SFT}_\text{TOPS}$  &  41.8\drop{0.4} & 10823.9 & \ddrop{8.4\%} & 30.7\bonus{2.5} & 11720.5 & \ddrop{33.7\%} & 82.7\bonus{0.9} & 6454.1 & \ddrop{17.5\%} & 31.8\drop{3.4} & 4666.3 & \bbonus{30.4\%} & 46.8\drop{0.1} & 8416.2 & \ddrop{12.3\%}\\

$\text{SimPO}_\text{DAST}$  &  22.9\drop{19.3} & 14029.4 & \bbonus{112.1\%} & 19.4\drop{8.9} & 11161.7 & \bbonus{70.0\%} & 68.1\drop{13.8} & 5955.5 & \bbonus{42.2\%} & 32.6\drop{2.6} & 5906.4 & \bbonus{2.1\%} & 35.7\drop{11.1} & 9263.3 & \bbonus{61.8\%}\\
$\text{SimPO}_\text{Shortest}$  &  35.8\drop{6.4} & 7422.1 & \bbonus{0.5\%} & 26.5\drop{1.8} & 7380.9 & \bbonus{0.1\%} & 77.7\drop{4.2} & 4020.8 & \ddrop{17.3\%} & 34.8\drop{0.4} & 3437.0 & \ddrop{51.5\%} & 43.7\drop{3.2} & 5565.2 & \ddrop{13.2\%}\\
$\text{SimPO}_\text{TOPS}$  &  15.6\drop{26.6} & 2647.4 & \bbonus{195.5\%} & 18.3\drop{9.9} & 2412.2 & \bbonus{81.3\%} & 58.5\drop{23.4} & 1497.8 & \bbonus{115.5\%} & 25.7\drop{9.5} & 1022.8 & \bbonus{129.4\%} & 29.5\drop{17.3} & 1895.1 & \bbonus{131.0\%}\\


\midrule
\multicolumn{16}{c}{\emph{DeepSeek-R1-Distill-Qwen-7B}} \\
\midrule

Base LRM  &  55.3 & 13062.1 & 1887.4 & 39.7 & 14241.9 & 1527.0 & 90.9 & 6177.3 & 1611.4 & 43.1 & 5575.8 & 855.7 & 57.2 & 9764.3 & 1470.4\\
\midrule

Vanilla RL  &  66.2 & 14264.7 & 1579.0 & 52.9 & 16305.3 & 1453.1 & 93.9 & 7259.8 & 1722.2 & 44.6 & 7300.7 & 945.2 & 64.4 & 11282.6 & 1424.9\\
\midrule

\rowcolor{bg!70} REO-RL (Exp) (ours)  &  64.0\drop{2.3} & 7671.5 & \ddrop{60.7\%} & 48.8\drop{4.2} & 8361.1 & \ddrop{57.3\%} & 93.4\drop{0.5} & 4144.6 & \ddrop{48.4\%} & 45.5\bonus{0.9} & 3687.0 & \ddrop{59.4\%} & 62.9\drop{1.5} & 5966.0 & \ddrop{55.9\%}\\
\rowcolor{bg!70} REO-RL (Oracle) (ours) &  63.9\drop{2.4} & 9348.6 & \ddrop{50.6\%} & 49.0\drop{4.0} & 9189.6 & \ddrop{66.5\%} & 94.7\bonus{0.8} & 4444.1 & \ddrop{57.9\%} & 45.5\bonus{0.9} & 4132.9 & \ddrop{57.8\%} & 63.3\drop{1.2} & 6778.8 & \ddrop{58.0\%}\\
\rowcolor{bg!70} REO-RL (Q-Spec) (ours)  &  63.9\drop{2.4} & 8407.8 & \ddrop{62.0\%} & 51.4\drop{1.6} & 9298.5 & \ddrop{75.3\%} & 93.4\drop{0.5} & 4230.6 & \ddrop{54.5\%} & 43.9\drop{0.7} & 3581.0 & \ddrop{38.9\%} & 63.1\drop{1.3} & 6379.4 & \ddrop{59.3\%}\\

RL w. Token Budget=1K  &  26.7\drop{39.6} & 1242.4 & \bbonus{176.4\%} & 19.7\drop{33.2} & 1168.0 & \bbonus{137.4\%} & 73.0\drop{20.9} & 960.4 & \bbonus{59.5\%} & 40.6\drop{3.9} & 690.7 & \ddrop{31.1\%} & 40.0\drop{24.4} & 1015.4 & \bbonus{98.0\%}\\
RL w. Token Budget=2K  &  36.5\drop{29.8} & 1929.6 & \bbonus{91.2\%} & 27.6\drop{25.3} & 1831.6 & \bbonus{61.0\%} & 84.3\drop{9.6} & 1427.5 & \ddrop{36.6\%} & 43.9\drop{0.7} & 1210.1 & \ddrop{74.8\%} & 48.1\drop{16.3} & 1599.7 & \bbonus{17.3\%}\\
RL w. Token Budget=4K  &  48.3\drop{17.9} & 3480.9 & \bbonus{0.8\%} & 35.4\drop{17.5} & 3345.2 & \ddrop{2.0\%} & 89.1\drop{4.8} & 2109.2 & \ddrop{59.7\%} & 43.5\drop{1.1} & 1772.8 & \ddrop{55.3\%} & 54.1\drop{10.3} & 2677.0 & \ddrop{27.5\%}\\

RL w. Len Group Norm. Rew.  &  64.2\drop{2.1} & 10608.1 & \ddrop{28.8\%} & 50.8\drop{2.1} & 11486.8 & \ddrop{47.9\%} & 94.5\bonus{0.5} & 4788.9 & \ddrop{43.7\%} & 45.2\bonus{0.6} & 4309.4 & \ddrop{45.5\%} & 63.7\drop{0.7} & 7798.3 & \ddrop{40.9\%}\\
RL w. Len-Harmonizing Rew.  &  65.5\drop{0.7} & 9167.7 & \ddrop{37.0\%} & 51.1\drop{1.8} & 10394.6 & \ddrop{46.1\%} & 94.5\bonus{0.5} & 4307.9 & \ddrop{52.7\%} & 44.4\drop{0.2} & 3617.9 & \ddrop{42.2\%} & 63.9\drop{0.5} & 6872.0 & \ddrop{44.9\%}\\
MRT  &  66.1\drop{0.1} & 9210.6 & \ddrop{42.4\%} & 50.5\drop{2.4} & 10559.3 & \ddrop{55.9\%} & 93.8\drop{0.2} & 4986.6 & \ddrop{38.6\%} & 44.7\bonus{0.1} & 4852.7 & \ddrop{32.4\%} & 63.8\drop{0.6} & 7402.3 & \ddrop{43.0\%}\\

$\text{SFT}_\text{Shortest}$  &  67.7\bonus{1.5} & 13197.2 & \ddrop{13.7\%} & 53.6\bonus{0.7} & 15208.6 & \ddrop{11.1\%} & 93.6\drop{0.3} & 6728.2 & \ddrop{4.0\%} & 44.6\drop{0.0} & 6623.5 & \ddrop{9.4\%} & 64.9\bonus{0.5} & 10439.4 & \ddrop{9.4\%}\\
$\text{SFT}_\text{TOPS}$  &  66.0\drop{0.2} & 13204.1 & \ddrop{9.0\%} & 52.0\drop{0.9} & 15170.8 & \ddrop{11.7\%} & 93.3\drop{0.6} & 6272.5 & \ddrop{11.9\%} & 44.2\drop{0.4} & 5759.7 & \ddrop{30.4\%} & 63.9\drop{0.6} & 10101.8 & \ddrop{14.1\%}\\

$\text{SimPO}_\text{DAST}$  &  60.1\drop{6.1} & 7500.5 & \ddrop{39.0\%} & 46.2\drop{6.7} & 7916.9 & \ddrop{47.9\%} & 92.0\drop{1.9} & 4051.7 & \ddrop{46.4\%} & 45.5\bonus{0.9} & 3402.6 & \ddrop{61.8\%} & 61.0\drop{3.4} & 5717.9 & \ddrop{47.3\%}\\
$\text{SimPO}_\text{Shortest}$  &  65.5\drop{0.7} & 9348.1 & \ddrop{38.5\%} & 50.8\drop{2.1} & 10292.3 & \ddrop{45.7\%} & 93.4\drop{0.5} & 4793.1 & \ddrop{36.2\%} & 45.4\bonus{0.8} & 4276.4 & \ddrop{46.7\%} & 63.8\drop{0.6} & 7177.5 & \ddrop{41.0\%}\\

HGPO & 68.1 \bonus{1.9} & 13556.8&\ddrop{4.6\%} & 53.1\bonus{0.2} & 15609.8&\bbonus{2.3\%} & 93.4\drop{0.5} & 7077.3&\bbonus{2.7\%} & 45.0\bonus{0.4} & 6873.4&\ddrop{15.6\%} & 64.9\bonus{0.5} & 10779.3&\ddrop{0.4\%} \\

\midrule
\multicolumn{16}{c}{\emph{Qwen3-4B}} \\
\midrule

Base LRM  &  72.1 & 14808.7 & 614.3 & 64.7 & 17618.7 & 335.0 & 97.0 & 8397.1 & 955.8 & 46.5 & 6864.3 & 654.3 & 70.1 & 11922.2 & 639.9\\
\midrule

\rowcolor{bg!70} REO-RL (Exp) (ours)  &  70.0\drop{2.1} & 10799.0 & \ddrop{87.8\%} & 59.9\drop{4.8} & 13063.4 & \ddrop{45.6\%} & 93.8\drop{3.2} & 5307.7 & \ddrop{74.4\%} & 45.2\drop{1.3} & 3653.3 & \ddrop{40.7\%} & 67.2\drop{2.8} & 8205.9 & \ddrop{65.2\%}\\
\rowcolor{bg!70} REO-RL (Q-Spec) (ours)  &  67.9\drop{4.2} & 12066.1 & \ddrop{21.7\%} & 62.1\drop{2.6} & 14755.0 & \ddrop{48.5\%} & 95.1\drop{2.0} & 5662.4 & \ddrop{63.6\%} & 46.0\drop{0.5} & 4203.5 & \ddrop{39.7\%} & 67.8\drop{2.3} & 9171.7 & \ddrop{45.5\%}\\
RL w. Token Budget=1K  &  25.9\drop{46.1} & 22992.1 & \bbonus{576.6\%} & 23.5\drop{41.1} & 21087.8 & \bbonus{815.4\%} & 73.6\drop{23.4} & 19078.4 & \bbonus{139.6\%} & 43.5\drop{3.0} & 10106.0 & \bbonus{27.9\%} & 41.6\drop{28.4} & 18316.1 & \bbonus{304.4\%}\\
RL w. Token Budget=2K  &  39.5\drop{32.6} & 12996.9 & \bbonus{311.4\%} & 30.1\drop{34.6} & 12546.4 & \bbonus{575.7\%} & 85.6\drop{11.4} & 7827.0 & \ddrop{0.8\%} & 46.0\drop{0.6} & 5542.0 & \ddrop{34.5\%} & 50.3\drop{19.8} & 9728.1 & \bbonus{141.0\%}\\
RL w. Token Budget=4K  &  52.6\drop{19.5} & 7909.6 & \bbonus{77.3\%} & 37.3\drop{27.4} & 8809.0 & \bbonus{370.7\%} & 90.8\drop{6.2} & 5316.3 & \ddrop{62.6\%} & 47.3\bonus{0.8} & 4302.1 & \ddrop{66.5\%} & 57.0\drop{13.1} & 6584.2 & \bbonus{26.7\%}\\
RL w. Token Budget=8K  &  64.1\drop{8.0} & 10240.9 & \ddrop{3.7\%} & 54.2\drop{10.5} & 11042.7 & \bbonus{64.7\%} & 93.0\drop{4.0} & 6050.0 & \ddrop{23.2\%} & 47.4\bonus{0.8} & 5067.9 & \ddrop{40.7\%} & 64.7\drop{5.4} & 8100.4 & \ddrop{11.5\%}\\
RL w. Len Group Norm. Rew.  &  70.0\drop{2.1} & 15022.9 & \bbonus{51.0\%} & 63.1\drop{1.6} & 16825.9 & \bbonus{104.7\%} & 95.6\drop{1.4} & 7682.7 & \ddrop{7.5\%} & 47.5\bonus{1.0} & 6322.9 & \ddrop{23.7\%} & 69.1\drop{1.0} & 11463.6 & \bbonus{17.1\%}\\
RL w. Len-Harmonizing Rew.  &  67.7\drop{4.4} & 15265.5 & \ddrop{0.4\%} & 56.2\drop{8.4} & 18466.5 & \bbonus{183.1\%} & 95.0\drop{2.0} & 7380.8 & \ddrop{41.9\%} & 45.5\drop{1.1} & 7771.8 & \bbonus{15.9\%} & 66.1\drop{4.0} & 12221.1 & \bbonus{12.3\%}\\
HGPO  &  72.4\bonus{0.3} & 15123.8 & \ddrop{16.4\%} & 65.4\bonus{0.7} & 17951.8 & \bbonus{49.6\%} & 96.2\drop{0.9} & 8858.2 & \bbonus{16.1\%} & 47.7\bonus{1.2} & 6797.4 & \ddrop{11.7\%} & 70.4\bonus{0.3} & 12182.8 & \bbonus{5.6\%}\\

\midrule
\multicolumn{16}{c}{\emph{Qwen3-8B}} \\
\midrule
Base LRM  &  75.1 & 15221.2 & 1402.2 & 67.2 & 17682.8 & 656.1 & 94.8 & 9050.7 & 1026.9 & 48.3 & 7281.1 & 682.0 & 71.3 & 12308.9 & 941.8\\
\midrule

\rowcolor{bg!70}  REO-RL (Exp) (ours)  &  73.6\drop{1.5} & 9285.4 & \ddrop{69.7\%} & 59.0\drop{8.2} & 11343.1 & \ddrop{68.2\%} & 95.1\bonus{0.3} & 4699.7 & \ddrop{98.6\%} & 47.8\drop{0.5} & 3671.8 & \ddrop{62.1\%} & 68.9\drop{2.5} & 7250.0 & \ddrop{75.9\%}\\
\rowcolor{bg!70} REO-RL (Q-Spec) (ours)  &  73.2\drop{1.9} & 10852.9 & \ddrop{37.1\%} & 64.4\drop{2.8} & 12645.2 & \ddrop{84.8\%} & 94.6\drop{0.2} & 5416.1 & \ddrop{71.5\%} & 46.8\drop{1.6} & 3952.7 & \ddrop{36.1\%} & 69.7\drop{1.6} & 8216.7 & \ddrop{54.6\%}\\

RL w. Token Budget=1K  &  32.3\drop{42.8} & 5170.2 & \bbonus{201.8\%} & 22.7\drop{44.5} & 5291.7 & \bbonus{439.9\%} & 67.6\drop{27.2} & 2042.4 & \bbonus{212.6\%} & 39.6\drop{8.8} & 1419.3 & \bbonus{77.6\%} & 40.5\drop{30.8} & 3480.9 & \bbonus{223.7\%}\\
RL w. Token Budget=2K  &  52.2\drop{22.9} & 5151.0 & \bbonus{33.1\%} & 35.1\drop{32.1} & 4696.1 & \bbonus{203.0\%} & 84.8\drop{9.9} & 3840.4 & \bbonus{8.8\%} & 47.4\drop{0.9} & 3034.1 & \ddrop{56.2\%} & 54.9\drop{16.5} & 4180.4 & \bbonus{39.9\%}\\
RL w. Token Budget=4K  &  62.3\drop{12.8} & 6905.4 & \ddrop{18.2\%} & 45.3\drop{21.9} & 6920.0 & \bbonus{49.2\%} & 89.0\drop{5.8} & 4274.1 & \ddrop{32.6\%} & 48.1\drop{0.2} & 3426.4 & \ddrop{62.0\%} & 61.2\drop{10.2} & 5381.5 & \ddrop{18.3\%}\\

RL w. Token Budget=8K  &  71.7\drop{3.4} & 7363.9 & \ddrop{84.8\%} & 51.4\drop{15.8} & 8359.4 & \bbonus{9.5\%} & 92.1\drop{2.7} & 4437.3 & \ddrop{73.4\%} & 48.4\bonus{0.0} & 3534.8 & \ddrop{74.2\%} & 65.9\drop{5.5} & 5923.8 & \ddrop{63.4\%}\\
RL w. Len Group Norm. Rew.  &  74.2\drop{0.9} & 12841.8 & \ddrop{20.2\%} & 64.8\drop{2.4} & 14991.5 & \ddrop{63.3\%} & 94.5\drop{0.2} & 6366.9 & \ddrop{48.0\%} & 46.7\drop{1.6} & 5138.1 & \ddrop{10.9\%} & 70.1\drop{1.3} & 9834.6 & \ddrop{33.6\%}\\
RL w. Len-Harmonizing Rew.  &  74.8\drop{0.3} & 15489.7 & \bbonus{17.7\%} & 65.8\drop{1.4} & 18779.6 & \bbonus{54.8\%} & 95.3\bonus{0.5} & 8655.2 & \ddrop{1.1\%} & 47.9\drop{0.5} & 7707.5 & \bbonus{17.2\%} & 71.0\drop{0.4} & 12658.0 & \bbonus{18.9\%}\\
HGPO &  75.3\bonus{0.2} & 15547.2 & \bbonus{10.6\%} & 68.0\bonus{0.8} & 18112.5 & \bbonus{7.6\%} & 95.3\bonus{0.5} & 9292.2 & \bbonus{9.7\%} & 48.9\bonus{0.5} & 7605.6 & \ddrop{5.9\%} & 71.9\bonus{0.5} & 12639.4 & \bbonus{6.8\%}\\
\bottomrule

\end{tabular}}
\caption{Accuracy, Generation Length, and Reasoning Efficiency Gap (REG) for all methods. For accuracy and REG, the relative changes compared to the vanilla RL baseline or the base LRMs are reported. {\method} could significantly reduce the gap from the LRM to the reasoning efficiency frontier with slight or even no accuracy drop at the same time. 
}
\label{tab:main-result}
\vspace{-4mm}
\end{table}

%% file: tables/frontier.tex
\begin{table}[htb]
\centering
\resizebox{0.99\textwidth}{!}{
\begin{tabular}{@{}c|ccccc|cc}
\toprule
&Claude Sonnet 3.7 (Thinking) & DeepSeek R1 & Qwen3-4B & Qwen3-8B & Qwen3-32B & Vanilla RL - 7B & {\method} (Exp) - 7B (ours) \\
\midrule
Length & 17478.87 & 5156.39 & 8631.61 & 8898.57 & 7755.99 & 7732.11& \textbf{4524.02} \\
Accuracy & 90.0\% & 98.2\% & 95.8\% & 94.9\% & 97.2\% & 93.5\% & 93.5\% \\
\bottomrule
\end{tabular}}
\caption{Comparing the response length with frontier LRMs.}
\label{tab:frontier}
\vspace{-6mm}
\end{table}

%% file: tables/ablation.tex
\begin{table}[htb]
\centering
\resizebox{1.0\textwidth}{!}{
\begin{tabular}{@{}c|ccc|ccc|ccc|ccc|ccc@{}}
\toprule
\multirow{2}{*}{Method} & \multicolumn{3}{c}{AIME 2024} & \multicolumn{3}{c}{AIME 2025} &\multicolumn{3}{c}{AMC 2023} & \multicolumn{3}{c}{Minerva Math} & \multicolumn{3}{c}{Average}\\
& Acc $\uparrow$ & Length $\downarrow$ & REG & Acc $\uparrow$ & Length $\downarrow$ & REG & Acc $\uparrow$ & Length $\downarrow$ & REG & Acc $\uparrow$ & Length $\downarrow$ & REG & Acc $\uparrow$ & Length $\downarrow$ & REG \\
\midrule
Base LRM  &  55.3 & 13062.1 & 1887.4 & 39.7 & 14241.9 & 1527.0 & 90.9 & 6177.3 & 1611.4 & 43.1 & 5575.8 & 855.7 & 57.2 & 9764.3 & 1470.4\\
\midrule
Vanilla RL  &  66.2 & 14264.7 & 1579.0 & 52.9 & 16305.3 & 1453.1 & 93.9 & 7259.8 & 1722.2 & 44.6 & 7300.7 & 945.2 & 64.4 & 11282.6 & 1424.9\\
\midrule
REO-RL (Exp)   &  64.0\drop{2.3} & 7671.5 & \ddrop{60.7\%} & 48.8\drop{4.2} & 8361.1 & \ddrop{57.3\%} & 93.4\drop{0.5} & 4144.6 & \ddrop{48.4\%} & 45.5\bonus{0.9} & 3687.0 & \ddrop{59.4\%} & 62.9\drop{1.5} & 5966.0 & \ddrop{55.9\%}\\
REO-RL (Oracle)  &  63.9\drop{2.4} & 9348.6 & \ddrop{50.6\%} & 49.0\drop{4.0} & 9189.6 & \ddrop{66.5\%} & 94.7\bonus{0.8} & 4444.1 & \ddrop{57.9\%} & 45.5\bonus{0.9} & 4132.9 & \ddrop{57.8\%} & 63.3\drop{1.2} & 6778.8 & \ddrop{58.0\%}\\
REO-RL (Task-Specific)   &  63.9\drop{2.4} & 8407.8 & \ddrop{62.0\%} & 51.4\drop{1.6} & 9298.5 & \ddrop{75.3\%} & 93.4\drop{0.5} & 4230.6 & \ddrop{54.5\%} & 43.9\drop{0.7} & 3581.0 & \ddrop{38.9\%} & 63.1\drop{1.3} & 6379.4 & \ddrop{59.3\%}\\
REO-RL (Task-Specific-Hard)  &  61.5\drop{4.8} & 6804.2 & \ddrop{53.5\%} & 47.9\drop{5.0} & 7092.5 & \ddrop{79.8\%} & 92.0\drop{2.0} & 3253.7 & \ddrop{60.7\%} & 44.9\bonus{0.3} & 2314.7 & \ddrop{72.8\%} & 61.5\drop{2.9} & 4866.3 & \ddrop{65.6\%}\\
REO-RL (Linear)  &  63.5\drop{2.7} & 8982.0 & \ddrop{46.1\%} & 49.9\drop{3.0} & 9001.6 & \ddrop{58.8\%} & 93.9$_{0.0}$ & 4290.9 & \ddrop{48.4\%} & 44.9\bonus{0.3} & 3733.6 & \ddrop{47.5\%} & 63.1\drop{1.3} & 6502.0 & \ddrop{50.3\%}\\
REO-RL (Exp) - K=10  &  64.0\drop{2.3} & 9352.7 & \ddrop{45.2\%} & 48.9\drop{4.1} & 9804.9 & \ddrop{47.8\%} & 94.1\bonus{0.2} & 4779.6 & \ddrop{52.0\%} & 44.9\bonus{0.3} & 4151.8 & \ddrop{46.3\%} & 63.0\drop{1.5} & 7022.3 & \ddrop{48.1\%}\\
REO-RL (Exp) - Coef=1  &  62.6\drop{3.6} & 7218.5 & \ddrop{70.0\%} & 47.2\drop{5.7} & 7525.9 & \ddrop{73.0\%} & 93.1\drop{0.8} & 3736.4 & \ddrop{63.4\%} & 45.4\bonus{0.8} & 3052.7 & \ddrop{73.4\%} & 62.1\drop{2.3} & 5383.4 & \ddrop{69.3\%}\\

\bottomrule
\end{tabular}}
\caption{Results of Ablation Study on DeepSeek-R1-Distilled-Qwen-7B on All Benchmarks}
\end{table}

%% file: sections/70_conclusion.tex
\section{Conclusion}
\label{sec:discuss}

In this work, we investigate efficient reasoning for LRMs. We introduce reasoning efficiency frontiers, which characterize the empirically optimal trade-off between response length and accuracy for LRMs. To quantify the reasoning efficiency of a fine-tuned LRM, we introduce the Reasoning Efficiency Gap (REG), a unified metric that captures both accuracy and length. 
We benchmark existing methods and reveal a substantial gap between current fine-tuning approaches and the frontiers. 
Our proposed method, {\method}, \gjx{could largely enhance the reasoning efficiency and become the closest to the optimal length-accuracy trade-off}. 

%% file: sections/appendix.tex



\section{Additional Discussion}
\label{app:discuss}

\paragraph{Connection to MRT~\citep{qi2025optimizinganytimereasoningbudget} and BRPO~\citep{qi2025optimizinganytimereasoningbudget}.} \gjx{Similar to our approach, {\method}, both MRT and BRPO share the idea of optimizing rewards under limited token budgets. However, key differences exist between {\method} and these prior or concurrent approaches. }

\gjx{First, the most important difference is the way to evaluate a partial response, especially the prompt and generation configuration to force the LRM to produce a plausible answer. {\method} requires the LRM to directly output the final answer with a generation budget and a prompt similar to ``The Final Answer is $\backslash$boxed\{''. MRT and BRPO employs a looser strategy that stops the thinking process with ``</think>'' and allows the model to produce a summarization within a moderate generation budget. Within the summarization phase, the model is able to perform lightweight thinking, which gives the model an additional try beyond the pre-specified token budget. Crucially, through RL training, this design has the risk of allowing the LRM to learn to perform \emph{budget-aware reasoning} within the summarization phase, which would be infeasible in practice since the optimal token budget for a question is unknown in advance, as also discussed in MRT~\citep{qu2025optimizing}. Therefore, we follow s1~\citep{muennighoff2025s1simpletesttimescaling} to force the LRM to produce a plausible answer with minimal additional reasoning efforts. }

\gjx{Second, the ways to select partial responses are different in all of these works. MRT relies on partitioning the model-generated responses into semantically complete steps through keywords. BRPO selects a small set of linearly spaced token budgets. {\method} introduces varying types of token budget selection strategies, such as exponentially spaced token budgets that achieve better results than the linear counterpart, and applying question-specific token budgets. Finally, MRT optimizes the single-step rewards while both BRPO and {\method} employ dense reward training and optimize total rewards across a wide range of token budgets.}


\section{Reproducibility}
\label{app:rep}

We will provide our code in the \href{https://github.com/samjia2000/Optimal-Reasoning-Efficiency}{https://github.com/samjia2000/Optimal-Reasoning-Efficiency}. Please refer to Sec.~\ref{app:frontier}  for the details on reasoning efficiency frontiers and reasoning efficiency gap, and Sec.~\ref{app:impl} for the implementation details.

\section{Implementation Details}
\label{app:impl}

\paragraph{Training Data.} For training data, we integrate data from DeepScaleR~\cite{deepscaler2025} and AReaL~\cite{areal2025}. For models with at least 4B sizes, we use the training data of AReaL-Boba-RL-7B~\citep{areal2025}. For 1.5B, we adopt the mixture of training data from DeepScaleR~\cite{deepscaler2025} and AReaL~\cite{areal2025} and remove duplicated problems. 

We implement the training algorithm with the AReaL framework~\citep{areal2025}, which supports SGLang~\citep{zheng2024sglangefficientexecutionstructured} for rollout generation. Below we detail implementation details of each training algorithm.

\paragraph{Online RL Training.}  We use PPO as the default online RL algorithm. Following standard practices in RL training for LLM reasoning~\citep{yu2025dapoopensourcellmreinforcement,hu2025openreasonerzeroopensourceapproach}, we do not utilize value model and KL regularization. The default training setting and hyperparameters for PPO training are listed in Tab.~\ref{tab:ppo-hyper}. 

For {\method} and baseline methods, we do not carry out RL training directly from the base LRM since it would lead to prolonged training and slower convergence. Instead, we perform further fine-tuning on the RL trained versions for both 1.5B and 7B settings. Specifically, we adopt AReaL-Boba-RL-1.5B~\citep{areal2025} and Skywork-OR1-Math-7B~\citep{skywork-or1-2025} as the starting points for further fine-tuning. Following the cluster configuration in Tab.~\ref{tab:ppo-hyper}, each experiment could finish within 4K GPU hours.

\begin{table}[ht]
\centering
\caption{Default training configurations and hyperparameters for PPO.}

\begin{tabular}{ll}
\toprule
\multicolumn{2}{l}{\textbf{Training Configuration}} \\
\midrule
Batch size (number of prompts) & 128 \\
Rollouts per prompt & 16 \\
Random seed & 1 \\
Cluster Config &  $8\times 8$ H800 (for 1.5B) / $16 \times 8$ H800 (others) \\
\midrule
\multicolumn{2}{l}{\textbf{PPO Parameters}} \\
\midrule
PPO Minibatches & 4 \\
Clipping $\epsilon$ & 0.2 \\
Advantage normalization  & True \\
Discount factor $\gamma$ & 1.0 \\
GAE $\lambda$ & 1.0 \\
Epochs & 2.0 \\
\midrule
\multicolumn{2}{l}{\textbf{Optimizer Parameters}} \\
\midrule
Optimizer & Adam \\
Learning rate & $5 \times 10^{-6}$ \\
Weight decay & 0.05 \\
$\beta_1$ & 0.9 \\
$\beta_2$ & 0.95 \\
Adam $\epsilon$ & $1 \times 10^{-5}$ \\
Gradient norm clipping & 1.0 \\
Learning rate scheduler & constant \\
Warmup steps proportion & 0.001 \\
\midrule
\multicolumn{2}{l}{\textbf{Generation Parameters}} \\
\midrule
Temperature & 1.0 \\
Top-p  & 1.0 \\
Top-k  & -1 \\
Max prompt length & 1024 \\
Min generation length & 0 \\
Max generation length & 24376 (for 1.5B) / 32768 (others) \\
\bottomrule
\end{tabular}
\label{tab:ppo-hyper}
\end{table} 

\paragraph{Supervised Fine-Tuning.} The default training configurations and hyperparameters for SFT are listed in Tab.~\ref{tab:sft-hyper}. 

\begin{table}[ht]
\centering
\caption{Default training configurations and hyperparameters for SFT.}

\begin{tabular}{ll}
\toprule
\multicolumn{2}{l}{\textbf{Training Configuration}} \\
\midrule
Batch size (number of prompt-answer pairs) & 512 \\
Cluster Config & $16 \times 8$ H800  \\
\midrule
\multicolumn{2}{l}{\textbf{SFT Parameters}} \\
\midrule
Epochs & 10 \\
Save Frequency Steps & 100 \\
use\_bf16 & True \\
Max Seq Length & 32768 \\
\midrule
\multicolumn{2}{l}{\textbf{Optimizer Parameters}} \\
\midrule
Optimizer & Adam \\
Learning rate & $1 \times 10^{-5}$ \\
Weight decay & 0.05 \\
$\beta_1$ & 0.9 \\
$\beta_2$ & 0.95 \\
Adam $\epsilon$ & $1 \times 10^{-5}$ \\
Gradient norm clipping & 1.0 \\
Learning rate scheduler & constant \\
Warmup steps proportion & 0.03 \\
\bottomrule
\end{tabular}
\label{tab:sft-hyper}
\end{table} 

\paragraph{Preference Learning.} We implement SimPO~\citep{meng2024simpo} in the AReaL framework~\citep{areal2025}. The default training configurations and hyperparameters for SimPO are listed in Tab.~\ref{tab:sft-hyper}.

\begin{table}[ht]
\centering
\caption{Default training configurations and hyperparameters for SimPO.}

\begin{tabular}{ll}
\toprule
\multicolumn{2}{l}{\textbf{Training Configuration}} \\
\midrule
Batch size (number of preference pairs) & 128 \\
Cluster Config & $16 \times 8$ H800  \\
\midrule
\multicolumn{2}{l}{\textbf{SimPO Parameters}} \\
\midrule
Epochs & 2 \\
Save Frequency Steps & 10 \\
use\_bf16 & True \\
Max Seq Length & 32768 \\
SimPO Coefficient $\beta$ & 1/2 \\
SimPO Coefficient $\gamma$ & 1.2/1.4 \\
\midrule
\multicolumn{2}{l}{\textbf{Optimizer Parameters}} \\
\midrule
Optimizer & Adam \\
Learning rate & $1\times 10^{-5}$ (for 1.5B) / $3 \times 10^{-6}$ (or 7B) \\
Weight decay & 0.05 \\
$\beta_1$ & 0.9 \\
$\beta_2$ & 0.95 \\
Adam $\epsilon$ & $1 \times 10^{-5}$ \\
Gradient norm clipping & 1.0 \\
Learning rate scheduler & constant \\
Warmup steps proportion & 0.03 \\
\bottomrule
\end{tabular}
\label{tab:simpo-hyper}
\end{table}

\section{Baselines} 
\label{app:baseline}

\paragraph{RL with Token Budgets.} In our online RL training with token budget constraints, we control the maximum generation length during each training phase. Rather than fine-tuning the LRM directly on a fixed token budget, we adopt a progressive length-shrinking strategy. We begin training with a 16K token budget. Once RL training at this level converges, we reduce the budget to 8K and continue training. This process is repeated, halving the token budget each time, until we reach the minimum budget of 512 tokens. This staged approach enables the model to gradually adapt to shorter generation lengths while maintaining reasoning performance.

\paragraph{RL with Length Rewards.} In the “RL with Length Group Normalized Rewards’' baseline~\citep{arora2025training}, for each question $x$ and the corresponding set of sampled responses $y^1,\cdots ,y^m$, the reward of response $y^i$ is computed as,
\begin{align*}
r(x,y^i)&=\mathbb I\{y^i\text{ is correct}\}(1-\alpha f(|y^i|))
\end{align*}
where the function $f$ normalizes $|y^i|$ according to the lengths of correct responses and applies a sigmoid function. Specifically,
\begin{align*}
f(|y^i|)=\sigma\left(\frac{|y^i|-\texttt{MEAN}(x)}{\texttt{STD}(x)}\right)
\end{align*}
where
\begin{align*}
\texttt{MEAN}(x)&=\mathbb E_{y\sim\pi(\cdot|x), s.t. y\text{ is correct}}[|y|]\\
\texttt{STD}(x)&=\sqrt{\mathrm{Var}_{y\sim\pi(\cdot|x), s.t. y\text{ is correct}}[|y|]}
\end{align*}

In the “RL with Length-Harmonizing Rewards'' baseline~\citep{luo2025o1}, for each question $x$ and the corresponding set of sampled responses $y^1,\cdots ,y^m$, the reward of response $y^i$ is computed as,
\begin{align*}
r(x, y^i)&=\frac{\overline L_{ref}(x)}{|y|} - 1 + \gamma\cdot (\mathbb I\{y\text{ is correct}\} - \overline A_{ref}(x))
\end{align*}
where $\overline L_{ref}(x)$ is the average response length of the reference model when taking $x$ as input and $\overline A_{ref}(x)$ is the average accuracy of the reference model. In our implementations, we set the models that serve as the starting point of RL training as the reference models.

In the MRT baseline~\citep{qu2025optimizing}, different from the original paper that only implements single-step optimization with offline collected response prefixes, we implement the online RL training version with dense rewards for MRT. For each question $x$ and the corresponding set of sampled responses , we partition each response into several steps $y=(y_1,\cdots, y_{s})$. In each training step, the model is updated by computing the policy gradient for the following objective,
\begin{align*}
\mathbb E_{x, y=(y_1,\cdots, y_{s})\sim\pi_{\theta'}(\cdot|x)}[&\sum_{i=1}^{s}\mathbb E_{y'_i\sim\pi_{\theta}(\cdot|x,y_{:i-1})}[\mathcal R(x, \text{Answer}(\pi_{\theta'},x,[y_{:i-1};y'])) \\
&- \mathcal R(x, \text{Answer}(\pi_{\theta'},x,y_{:i-1}))+\alpha\cdot \mathcal R(x, y)]]
\end{align*}
where $\alpha$ is the weight for the overall accuracy and is set as $0.2$. 

\paragraph{Hybrid Reasoning.} \gjx{We use ``$\backslash n$'' and ``$\backslash n$$\backslash n$</think>'' as the tokens to start thinking and no-thinking. We first run SFT over model-generated responses in the thinking and no-thinking modes to ensure the model initially has a non-zero probability for no-thinking modes. Then we apply HGPO training with a margin of $0.2$ following ~\citep{jiang2025thinkneedlargehybridreasoning}.}

\paragraph{Supervised Fine-Tuning.} In $\text{SFT}_{\text{Shortest}}$, we generate $16$ outputs for each question in the training dataset. Then we select the correct response with the shortest length for each question to construct the SFT dataset. In $\text{SFT}_{\text{TOPS}}$, we follow \citep{yang2025thinkingoptimalscalingtesttimecompute} to prompt the LRM to generate responses with three different types of reasoning efforts. We strictly follow the prompts used in \citep{yang2025thinkingoptimalscalingtesttimecompute}. For each type of reasoning effort, we generate $16$ responses. To construct the SFT dataset, the shortest correct response among all $48$ responses are gathered.

\paragraph{Preference Learning.} We adopt three strategies for constructing the preference datasets. In $\text{SimPO}_{\text{Shortest}}$, we adopt the responses generated for  $\text{SFT}_{\text{Shortest}}$ and select the shortest correct response and the longest response as the preference pair for each question. In $\text{SimPO}_{\text{TOPS}}$, we use the same preference construction strategy as $\text{SimPO}_{\text{Shortest}}$ but on the responses generated through $\text{TOPS}$, which contain reasoning traces with different reasoning efforts. Finally, in $\text{SimPO}_{\text{DAST}}$, we again utilize the responses generated for  $\text{SFT}_{\text{Shortest}}$ but adopt a different preference construction strategy. In the preference dataset of $\text{SimPO}_{\text{DAST}}$, each pair falls into one of the two cases: it either contains two correct responses where the positive sample is much shorter than the negative sample, or contains two incorrect responses where the positive sample is much longer than the negative sample.

\paragraph{Other Methods We Have Tried.} We have also experimented with Z1~\citep{yu2025z1efficienttesttimescaling} that constructs code-based reasoning traces, and L1~\citep{aggarwal2025l1} that fine-tunes the LRMs to follow token budget instruction with RL. We find Z1 having poor performance on mathematical reasoning tasks, demonstrating significantly lower accuracy to that of the base LRMs as illustrated in Tab.~\ref{tab:z1}. For L1, we find L1 mainly works under tight token budgets, i.e. less than $4K$. When we extend the training approach of L1 to a larger context length, i.e. 24K for 1.5B models, we find it hard to make the LRM learning to follow strict token budget instructions through RL. Consequently, the resulted models, L1-Exact-24K-1.5B and L1-Max-24K-1.5B could not follow the token budget instruction, as shown in Tab,~\ref{tab:l1}.

\begin{table}[h]
\centering
    \resizebox{0.8\textwidth}{!}{
\begin{tabular}{@{}c|cccccc@{}}
\toprule
\multirow{2}{*}{Method} & \multicolumn{2}{c}{AIME24} & \multicolumn{2}{c}{MATH500} & \multicolumn{2}{c}{GPQA} \\
 & Accuracy (\%) & Length  & Accuracy (\%) & Length & Accuracy (\%) & Length  \\
\midrule 
Base LRM & 31.5 & 16747.6 & 83.6 & 5633.1 & 44.6 & 10325.2\\
\midrule
Z1  & 10.0 & 15106.0 & 63.6 & 4904.1 & 61.6 & 9004.1  \\
\bottomrule
\end{tabular}}
\caption{Result of Z1 on DeepSeek-R1-Distill-Qwen-1.5B.~\citep{yu2025z1efficienttesttimescaling}}
\label{tab:z1}
\end{table}

\begin{table}[h]
\centering
    \resizebox{1.0\textwidth}{!}{
\begin{tabular}{@{}c|cccccc@{}}
\toprule
\multirow{2}{*}{Instructed Token Budget} & \multicolumn{2}{c}{AMC23} & \multicolumn{2}{c}{AIME24} & \multicolumn{2}{c}{AIME25} \\
 & Accuracy (\%) & Length  & Accuracy (\%) & Length & Accuracy (\%) & Length  \\
\midrule 
2048 & 63.9 & 20267.7 & 38.6 & 18067.4 & 27.2 & 19657.7 \\
4096 & 63.5 & 20142.6 & 38.4 & 17961.3 & 26.9 & 19711.1 \\
8192 & 62.9 & 20191.6 & 38.4 & 17868.7 & 26.8 & 19673.5 \\
\bottomrule
\end{tabular}}
\caption{Result of L1-Exact-24K-1.5B.}
\label{tab:l1}
\end{table}

\section{{\method}}
\label{app:method}

\subsection{Implementing {\method}}

\paragraph{Generation Phase.} In the generation phase of {\method}, there are two rounds of LRM generation. In the first round, multiple responses are generated for each question in the training batch. In the second round, to compute the length-constrained rewards for each of the responses and across all selected token budgets, we choose all truncated responses $y_{:L_i}$ and apply a prompt to enforce the LRM to generate the final answer given incomplete reasoning traces, i.e. $a=\pi_\theta(\cdot|x, y_{:L},[\text{The Final Answer is}]).$ We follow the prompt employed by \citep{fu2024efficiently} and \citep{fu2025reasoning}.

\begin{tcolorbox}[colback=blue!5!white,colframe=blue!75!black,title=Prompt for Forcing LRM to Produce Answer]
...
\newline
Oh, I suddenly got the answer to the whole problem. **Final Answer**:
\newline
[$\backslash$boxed\{
\end{tcolorbox}

\paragraph{Dense Reward RL.} {\method} obtains dense rewards through forcing the LRM to generate answer under various token budgets. The objective of {\method} is as follows,
\begin{align*}
\textbf{REO-RL:}&&\mathscr{L}_{{\method}}(\theta,\mathcal D)= \mathbb E_{x\sim \mathcal D}\left[\mathbb E_{y\sim \pi_\theta(\cdot|x)}\left[\sum_{i=1}^{N+1}c_i\mathcal{R}(x,\text{Answer}(\pi_\theta, x, y_{:L_i}))\right]\right]
\label{eq:reo-rl}
\end{align*}
where $c_i=\frac{L_{i+1}-L_{i-1}}{2}$ for $1\le i\le N$ 
and $c_{N+1}=\frac{L_{\text{max}}-L_N}{2}$ are the coefficient for the $i$-th token budget.

To perform policy update, we compute the return for each section between two consecutive token budgets $L_i$ and $L_{i+1}$. For $1\le i \le N$, we compute,
\begin{align*}
\mathrm{Return_i(x,y)}=\sum_{j=i}^{N+1}c_j\mathcal R(x,\text{Answer}(\pi_\theta,x,y_{:L_j}))
\end{align*}

Since we disable value model in the training process, the computed returns are then used directly as the advantages for loss computation. 
\gjx{In practice, for each prompt, we apply group normalization for each $i\in[1, N]$ before the policy update step, i.e. $\mathrm{Adv}_i(x,y^j)=\frac{\mathrm{
Return
(x,y^j)}-\frac{1}{M}\sum_{j'}\mathrm{Return}(x,y^{j'})}{\sqrt{\mathrm{Var}_{j'}[\mathrm{Return}(x,y^{j'})]}}$. Then $\text{Adv}_i(x,y)$ will be used as advantages to update the tokens $y_{L_{i}:L_{i+1}}$.}

\subsection{REO-RL (Q-Spec)}

\paragraph{REO-RL (Q-Spec).} \gjx{Alternatively, we note that there should exist a specific minimum token budget $L^x$ for each question $x$. This minimum token budget reflects the lowest token budget under which the base LRM could learn to achieve the same accuracy as the well-trained counterpart under the full token budget. }

\gjx{To derive the question-specific minimum token budgets, we take a practical estimation approach that extracts from the rollouts generated during the training process of various configurations in Sec.~\ref{sec:estimating-optimal}. For a question $x$, the RL experiments in Sec.~\ref{sec:estimating-optimal} produce a large amount of rollouts generated by different fine-tuned models associated with the corresponding length-constrained rewards, i.e. $\mathcal D_{\text{rollouts}}=\{(\hat\theta, x, y^1,\cdots, y^M,r(x,\cdot;\hat\theta))\}$ where $\hat \theta$ denotes a fine-tuned LRM and $y^1,\cdots, y^M$ are $M$ rollouts generated by $\hat \theta$ for question $x$. From $\mathcal D_{\text{rollouts}}$, we compute $L^x$ that satisfies,}

\begin{align}
L^x=\min\left\{L|L\in[1,L_{\text{max}}]\text{ s.t. }\max_{(\hat\theta, x, y^1,\cdots, y^M)\in\mathcal D_{\text{rollout}}} \frac{\sum_i r(x,y^i_{:L};\hat\theta)}{M}\ge\max_{(\hat\theta, x, y^1,\cdots, y^M)\in\mathcal D_{\text{rollout}}} \frac{\sum_i r(x,y^i_{:L_{\text{max}}};\hat\theta)}{M}\right\}
\end{align}

\gjx{In practice, evaluating $r(x,y^i_{:L};\hat\theta)$ for all history models $\hat\theta$, history rollouts and all possible $L$ could be expensive due to additional inference runs to evaluated truncated responses as in Eq.~\ref{eq:answer-forcing}. When computing the minimum token budgets, we directly evaluate the correctness of the partial response $y^i_{:L}$ without the additional operation to produce the answer.}

\gjx{In {\method} (Question-Specific), we set $N=1$ and $L_1=L^x$ for each question $x$, leading to the objective,}
\begin{align}
\textbf{REO-RL (Q-Spec):}&&\mathscr{L}_{{\method}\text{ (Q-Spec)}}(\theta,\mathcal D)= \mathbb E_{x\sim \mathcal D}\left[\mathbb E_{y\sim \pi_\theta(\cdot|x)}\left[c_1\cdot r(x,y_{:L^x};\theta)+c_2\cdot r(x,y_{:L_{\text{max}};\theta})\right]\right]
\label{eq:reo-rl}
\end{align}

\gjx{where $c_1=\frac{L^x}{2},c_2=\frac{L_{\text{max}}-L^x}{2}$ following Eq.~\ref{eq:reo-rl}.}

\section{Reasoning Efficiency Frontiers \& Reasoning Efficiency Gap}
\label{app:frontier}

\paragraph{Reasoning Efficiency Frontiers.} The two boxes below record the detailed lengths and accuracies for points on the estimated reasoning efficiency frontiers for DeepSeek-R1-Distill-Qwen-1.5B and DeepSeek-R1-Distill-Qwen-7B, respectively.

\begin{tcolorbox}
[colback=blue!5!white,colframe=blue!75!black,title=Reasoning Efficiency Frontier for Qwen3-4B]
    x, y = [0, 64, 128, 192, 256, 320, 384, 448, 512, 576, 640, 704, 768, 832, 896, 960, 1024, 2048, 3072, 4096, 5120, 6144, 7168, 8192, 9216, 10240, 11264, 12288, 13312, 14336, 15360, 16384, 17408, 18432, 19456, 20480, 21504, 22528, 23552, 24576, 25600, 26624, 27648, 28672, 29696, 30720, 31744, 32768], [0.05864545036764705, 0.07844286151960785, 0.08253484987745098, 0.09157475490196078, 0.09366000306372549, 0.10081380208333333, 0.10913181678921569, 0.12006548713235293, 0.14239813112745098, 0.16250957414215686, 0.18545879289215683, 0.20630935968137254, 0.22248008578431372, 0.23359183517156862, 0.24719860600490196, 0.25932329963235295, 0.26839958639705885, 0.37697610294117645, 0.4535807291666667, 0.5027018229166667, 0.5349207261029412, 0.5595760569852941, 0.5795496323529412, 0.5964384191176471, 0.6082050398284314, 0.6200099571078431, 0.6309244791666666, 0.6428308823529412, 0.6487189797794117, 0.6562059589460785, 0.6632372089460784, 0.6686408547794117, 0.6734145220588235, 0.6795764399509804, 0.6835171568627451, 0.6894416360294119, 0.693035768995098, 0.6970358455882353, 0.7003484987745099, 0.7034734987745098, 0.7032781862745098, 0.7061714920343137, 0.7074371936274509, 0.7081533394607843, 0.7090647977941177, 0.7103381587009804, 0.7101064644607843, 0.707257199754902]
\end{tcolorbox}

\begin{tcolorbox}
[colback=blue!5!white,colframe=blue!75!black,title=Reasoning Efficiency Frontier for Qwen3-8B]
    x, y = [0, 64, 128, 192, 256, 320, 384, 448, 512, 576, 640, 704, 768, 832, 896, 960, 1024, 2048, 3072, 4096, 5120, 6144, 7168, 8192, 9216, 10240, 11264, 12288, 13312, 14336, 15360, 16384, 17408, 18432, 19456, 20480, 21504, 22528, 23552, 24576, 25600, 26624, 27648, 28672, 29696, 30720, 31744, 32768], [0.09000842524509803, 0.10032935049019608, 0.09675245098039215, 0.11996208639705883, 0.1271771599264706, 0.1278818167892157, 0.14018267463235295, 0.16002795649509804, 0.1714862898284314, 0.2006414675245098, 0.21497970281862744, 0.24749348958333334, 0.2608475030637255, 0.2683632046568627, 0.2772250306372549, 0.28248697916666665, 0.29307789522058825, 0.39007352941176476, 0.4767539828431373, 0.5257927389705882, 0.5632065716911765, 0.5868183210784313, 0.6074180453431373, 0.6239028033088235, 0.6389916513480391, 0.6473824295343137, 0.6582471660539215, 0.666616881127451, 0.6744542738970589, 0.6808746936274509, 0.6869829963235294, 0.6924479166666666, 0.695980775122549, 0.7023954503676471, 0.7065142463235294, 0.7109911151960785, 0.7135799632352942, 0.7167777267156862, 0.7191789215686275, 0.7239296109068628, 0.726351868872549, 0.7280005361519608, 0.7298598345588236, 0.7314587162990196, 0.7330135569852941, 0.7345473345588236, 0.7354587928921568, 0.7288985906862746]
\end{tcolorbox}

\begin{tcolorbox}
[colback=blue!5!white,colframe=blue!75!black,title=Reasoning Efficiency Frontier for DeepSeek-R1-Distill-Qwen-1.5B]
x, y = [0, 64, 128, 192, 256, 320, 384, 448, 512, 576, 640, 704, 768, 832, 896, 960, 1024, 2048, 3072, 4096, 5120, 6144, 7168, 8192, 9216, 10240, 11264, 12288, 13312, 14336, 15360, 16384, 17408, 18432, 19456, 20480, 21504, 22528, 23552, 24576, 25600, 26624, 27648, 28672, 29696, 30720, 31744, 32768], [0.06101600796568627, 0.05281671262254902, 0.06075367647058823, 0.07656441482843138, 0.09422679227941178, 0.09907322303921567, 0.11950635723039214, 0.14104051776960785, 0.16222426470588236, 0.17982153799019607, 0.19957299325980393, 0.22226179534313725, 0.23991076899509806, 0.25617723651960783, 0.2744064031862745, 0.2836722579656863, 0.29287109375, 0.3738262101715686, 0.413882506127451, 0.44010225183823526, 0.4490253523284314, 0.4542604932598039, 0.4631778492647059, 0.4701803768382353, 0.4771963082107843, 0.48008961397058825, 0.48356885723039217, 0.48675130208333334, 0.4884727328431372, 0.48968098958333334, 0.49039713541666663, 0.4909466911764706, 0.4908088235294118, 0.4905771292892157, 0.4908662683823529, 0.49117647058823527, 0.4919289981617647, 0.4914445465686274, 0.4919002757352941, 0.4920955882352941, 0.4919002757352941, 0.4919289981617647, 0.4919289981617647, 0.4919577205882353, 0.4919002757352941, 0.4919289981617647, 0.4920955882352941, 0.49215303308823527]
\end{tcolorbox}

\begin{tcolorbox}
[colback=blue!5!white,colframe=blue!75!black,title=Reasoning Efficiency Frontier for DeepSeek-R1-Distill-Qwen-7B]
x, y = [0, 64, 128, 192, 256, 320, 384, 448, 512, 576, 640, 704, 768, 832, 896, 960, 1024, 2048, 3072, 4096, 5120, 6144, 7168, 8192, 9216, 10240, 11264, 12288, 13312, 14336, 15360, 16384, 17408, 18432, 19456, 20480, 21504, 22528, 23552, 24576, 25600, 26624, 27648, 28672, 29696, 30720, 31744, 32768], [0.06838809742647059, 0.07453469669117647, 0.08459520526960784, 0.09630629595588236, 0.09808900122549019, 0.11145450367647058, 0.14210707720588237, 0.16989123774509804, 0.1799383425245098, 0.2068627450980392, 0.22867455575980392, 0.24759880514705881, 0.26799938725490197, 0.2935891544117647, 0.31299019607843137, 0.33319546568627456, 0.38615196078431374, 0.4691272212009804, 0.50302734375, 0.5375919117647059, 0.5522518382352941, 0.5640356924019608, 0.5759803921568628, 0.5929974724264706, 0.6051547181372549, 0.61650390625, 0.6240272671568627, 0.6303423713235294, 0.63447265625, 0.63720703125, 0.6392252604166667, 0.64111328125, 0.64345703125, 0.6455403645833333, 0.6470377604166666, 0.6482747395833333, 0.6487955729166667, 0.6512044270833334, 0.6522460937500001, 0.6525065104166667, 0.6532877604166667, 0.6522460937500001, 0.6527669270833334, 0.6535481770833333, 0.6535481770833333, 0.6535481770833333, 0.6532877604166667, 0.6532877604166667]
\end{tcolorbox}

\paragraph{Evaluating Reasoning Efficiency Gap.} To practically evaluate REG, instead of strictly following Eq.~\ref{eq:reg}, we obtain approximations of $\sum_{L=1}^{L_{\text{max}}} \hat J_{\text{optimal}}$ and $\sum_{L=1}^{L_{\text{max}}}J(\mathcal D,\theta, L)$ through numerical integration on a set of token budgets respectively, in a similar way to Eq.~\ref{eq:approx-optimal-max}. We select $\{L_1,\cdots,L_N\}=\{64i|0\le i<16\}\cup\{1024i|1\le i\le 16\}$. Note that we set $L_{\text{max}}=16K$ instead of $L_{\text{max}}=32K$ to focus on the efficiency gap under lower token budgets.